\documentclass[11pt, letterpaper, logo, onecolumn, copyright]{main}

\renewcommand{\today}{2026-9-5}

\usepackage[authoryear,sort&compress,round]{natbib}
\usepackage[inkscapeformat=png]{svg}
\usepackage[most,breakable,skins]{tcolorbox}
\tcbuselibrary{skins}
\usepackage{lipsum}
\usepackage{tabularx}
\usepackage{afterpage}
\usepackage{booktabs}
\usepackage{subcaption}
\usepackage{makecell}
\usepackage{multirow}
\usepackage{bm}
\usepackage{multicol}
\usepackage{array}
\usepackage{float}
\usepackage{listings}
\usepackage{fontawesome5}
\usepackage{hyperref}
\usepackage{amssymb,graphicx}
\usepackage[dvipsnames]{xcolor}
\usepackage{cleveref}
\usepackage{longtable}
\usepackage{pdflscape}
\usepackage{adjustbox}
\usepackage{nicematrix}
\usepackage{CJKutf8}
\usepackage{ragged2e}
\usepackage{colortbl}
\usepackage{enumitem}
\usepackage{algorithm}
\usepackage{algorithmic}
\usepackage{pifont}
\usepackage{amsmath}
\usepackage{amsthm}
\usepackage{circledsteps}
\usepackage{diagbox}
\usepackage{bookmark}
\usepackage{wrapfig}
\usepackage{placeins}
\usepackage{needspace}
\usepackage{xspace}
\usepackage{tikz}
\usepackage[normalem]{ulem}
\usepackage{pgfplots}
\usepackage{pgfplotstable}
\usepackage{docmute}

\pgfplotsset{compat=1.18}
\BeforeBeginEnvironment{tabular}{\begin{adjustbox}{max width=\linewidth}}
\AfterEndEnvironment{tabular}{\end{adjustbox}}

\let\ArxivOriginalFigure\figure
\let\endArxivOriginalFigure\endfigure
\let\ArxivOriginalWrapFigure\wrapfigure
\let\endArxivOriginalWrapFigure\endwrapfigure
\newcounter{ArxivOrdinaryFigure}
\newif\ifArxivIntroFigure
\RenewDocumentEnvironment{figure}{O{tbp}}{%
  \stepcounter{ArxivOrdinaryFigure}%
  \ifnum\value{ArxivOrdinaryFigure}=1
    \ArxivIntroFiguretrue
    \ArxivOriginalWrapFigure{R}{0.48\textwidth}%
    \setlength{\columnwidth}{\linewidth}%
    \centering
    \captionsetup{width=\linewidth}%
  \else
    \ArxivIntroFigurefalse
    \ArxivOriginalFigure[#1]%
    \centering
    \captionsetup{width=\linewidth}%
  \fi
}{%
  \ifArxivIntroFigure
    \endArxivOriginalWrapFigure
  \else
    \endArxivOriginalFigure
  \fi
}

\let\ArxivOriginalTable\table
\let\endArxivOriginalTable\endtable
\RenewDocumentEnvironment{table}{O{tbp}}{%
  \ArxivOriginalTable[#1]%
  \centering
  \begin{minipage}{0.50\textwidth}%
  \centering
  \captionsetup{width=\linewidth}%
}{%
  \end{minipage}%
  \endArxivOriginalTable
}

\definecolor{medgray55}{gray}{0.55}
\definecolor{medgray}{gray}{0.7}
\definecolor{litegray}{gray}{0.9}
\definecolor{gblue}{RGB}{210,227,252}
\definecolor{gred}{RGB}{250,210,207}
\definecolor{gyellow}{RGB}{254,239,195}
\definecolor{ggreen}{RGB}{206,234,214}
\definecolor{gorange}{RGB}{254,223,200}
\definecolor{gblue9}{RGB}{23,78,166}
\definecolor{gred9}{RGB}{165,14,14}
\definecolor{gyellow9}{RGB}{227,116,0}
\definecolor{ggreen9}{RGB}{13,101,45}
\definecolor{gorange9}{RGB}{176,96,0}
\definecolor{myblue}{rgb}{0,0,1}
\definecolor{myred}{rgb}{1,0,0}
\definecolor{mylightgray}{gray}{0.95}
\definecolor{myCite}{HTML}{1C4587}
\definecolor{highlightblue}{HTML}{185ABC}
\definecolor{cellHighlight}{HTML}{dbefff}

\newcolumntype{L}[1]{>{\raggedright\let\newline\\\arraybackslash\hspace{0pt}}m{#1}}
\newcolumntype{C}[1]{>{\centering\arraybackslash}m{#1}}
\newcolumntype{R}[1]{>{\raggedleft\let\newline\\\arraybackslash\hspace{0pt}}m{#1}}

\usepackage{minitoc}

\noptcrule

\newtcolorbox{insightbox}[2][]{%
  colback=gray!5,
  colframe=gray!60,
  fonttitle=\small\bfseries,
  title={#2},
  boxrule=0.5pt,
  arc=2pt,
  left=4pt,right=4pt,top=2pt,bottom=2pt,
  breakable,
  #1
}

\let\cite\citep
\hypersetup{
  citecolor=myCite,
  linkcolor=myCite,
  urlcolor=myCite
}

\title{Strategy Accumulation and Guided Execution for Automated LLM Fine-Tuning}

\author{
\textbf{Haoran Zhao}$^{1,2*}$, \textbf{Wei Du}$^{2*}$, \textbf{Dingwen Yang}$^{1*}$, \textbf{Jixuan Huang}$^1$, \textbf{Junlin Shang}$^1$, \\
\textbf{Lingyong Fang}$^{3,2}$, \textbf{Ya Guo}$^{2\dag}$, \textbf{Tao Gui}$^{1\dag}$, \textbf{Qi Zhang}$^{1}$, \textbf{Xuanjing Huang}$^{1}$
\\
$^1$Fudan University \quad $^2$Ant Group \quad $^3$Shanghai Jiaotong University\\
\texttt{hrzhao26@m.fudan.edu.cn, guoya.gy@antgroup.com, tgui@fudan.edu.cn}
}

\begin{abstract}
Producing task-specific large language models requires discovering effective training strategies through experimentation. Automated fine-tuning systems have made this experimentation feasible with far less manual effort. However, these systems are stateless: each search discards its discovered strategies, dataset insights, and hyperparameter findings once it ends. Every new task must then repeat this costly search from a cold start. To address this, we propose \textit{\textbf{S}trategy \textbf{A}ccumulation and \textbf{G}uided \textbf{E}xecution (\textbf{SAGE})}, a two-stage framework that makes automated fine-tuning search cumulative. In the first stage, a multi-agent pipeline performs Monte Carlo Tree Search-based exploration. A parallel Distillation Agent extracts task-specific exploration records and confidence-scored cross-task insights, which together constitute a structured experience repository. In the second stage, SAGE retrieves relevant experience from this repository and selects what applies to guide training on the new task. We evaluate SAGE on nine unseen tasks spanning both single- and cross-category settings. In single-round execution, SAGE's accumulated experience raises the average relative improvement over baseline from 3.2\% to 15.6\%---a 12.4-percentage-point gain over the same pipeline without it. These results show that persistent strategy experience provides effective guidance for automated fine-tuning on unseen tasks.
\end{abstract}

\begin{document}

\doparttoc
\faketableofcontents

\begingroup
  \renewcommand\thefootnote{}
  \footnote{\textsuperscript{*}Equal contribution. \textsuperscript{\dag}Corresponding authors.}
  \addtocounter{footnote}{-1}
\endgroup

\maketitle

\let\maketitle\relax
\renewenvironment{abstract}{\setbox0=\vbox\bgroup}{\egroup}

\maketitle

\begin{abstract}
Producing task-specific large language models requires discovering effective training strategies through experimentation. Automated fine-tuning systems have made this experimentation feasible with far less manual effort. However, these systems are stateless: each search discards its discovered strategies, dataset insights, and hyperparameter findings once it ends. Every new task must then repeat this costly search from a cold start. To address this, we propose \textit{\textbf{S}trategy \textbf{A}ccumulation and \textbf{G}uided \textbf{E}xecution (\textbf{SAGE})}, a two-stage framework that makes automated fine-tuning search cumulative. In the first stage, a multi-agent pipeline performs Monte Carlo Tree Search-based exploration. A parallel Distillation Agent extracts task-specific exploration records and confidence-scored cross-task insights, which together constitute a structured experience repository. In the second stage, SAGE retrieves relevant experience from this repository and selects what applies to guide training on the new task. We evaluate SAGE on nine unseen tasks spanning both single- and cross-category settings. In single-round execution, SAGE's accumulated experience raises the average relative improvement over baseline from 3.2\% to 15.6\%---a 12.4-percentage-point gain over the same pipeline without it. These results show that persistent strategy experience provides effective guidance for automated fine-tuning on unseen tasks.
\end{abstract}

\section{Introduction}

Fine-tuning has established large language models (LLMs) as central components across vertical domains~\cite{li2023chatdoctor, roziere2023code}. However, finding an effective fine-tuning scheme requires searching a high-dimensional strategy space spanned by the training algorithm, dataset construction, data format, and hyperparameter configuration. The interactions among these dimensions are difficult to predict. The choice of training method constrains the optimal hyperparameter range, and the data distribution in turn shapes the design of the data format. As a result, the fine-tuning process depends heavily on the experience and judgment of domain experts, together with repeated experimental iteration, typically consuming days to weeks of effort. This reliance on expertise and manual labor becomes a key bottleneck to scaling.

\begin{figure}[t]
\centering
\includegraphics[width=0.95\columnwidth]{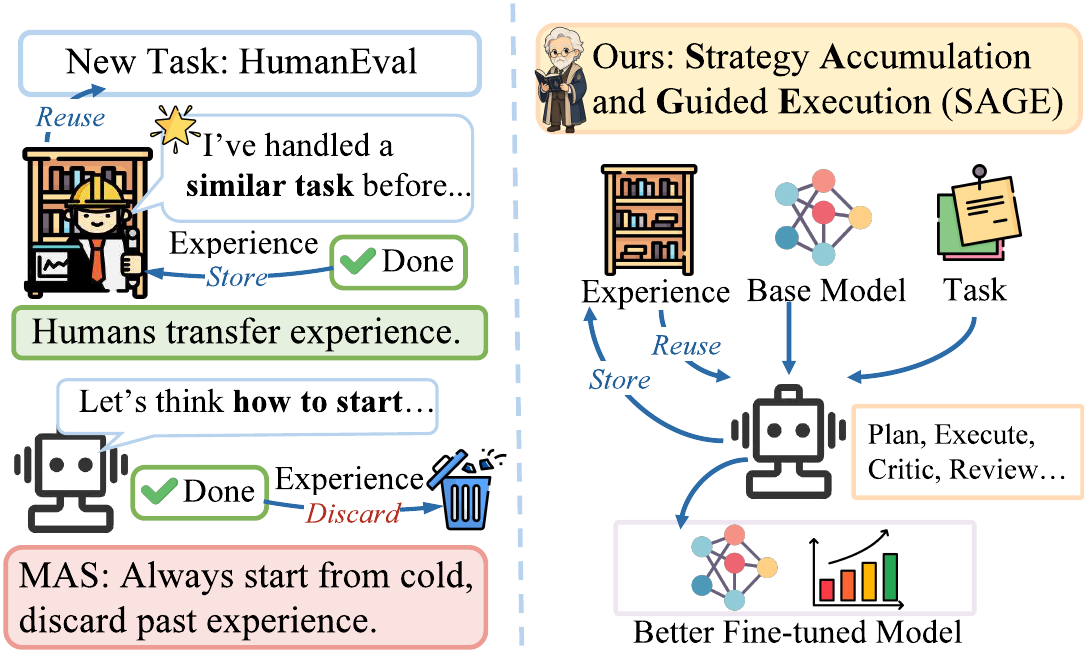}
\caption{Overview of SAGE. Conventional systems discard past experience; SAGE instead retrieves and updates accumulated experience to guide fine-tuning on new tasks.}
\label{fig:intro}
\end{figure}

The development of AI-driven automated research systems has made it feasible to automate experimental design, execution, and iterative refinement in machine learning~\cite{DBLP:journals/corr/abs-2502-13138, DBLP:journals/corr/abs-2506-16499, DBLP:journals/corr/abs-2604-13018}. TREX~\cite{DBLP:journals/corr/abs-2604-14116} extends this paradigm to LLM fine-tuning. It uses a Monte Carlo Tree Search (MCTS)-driven multi-agent system to iteratively refine training schemes, achieving results competitive with manually engineered baselines on multiple tasks. However, such fine-tuning systems remain costly, requiring dozens of full rounds of post-training for each task. Because the accumulated knowledge is discarded once a task is solved, even a highly similar task pays this cost again.

A natural question then arises:
\textbf{\textit{Can knowledge acquired from past fine-tuning searches be accumulated and transferred to accelerate search on new tasks?}} Human experts approach related problems by drawing on reusable strategic knowledge---what works, what fails, and under what conditions---rather than treating each task as a cold start~\cite{ball1997problem,arab2022exploratory}. Recent work in automated ML engineering provides preliminary evidence for this principle: HASTE distills experimental outcomes into scope-aware skills and transfers them across Kaggle tasks, reducing refinement effort in warm start~\cite{kim2026solve}. Yet how to accumulate and selectively reuse such knowledge in costly LLM fine-tuning searches remains underexplored.

To address this gap, we propose \textbf{S}trategy \textbf{A}ccumulation and \textbf{G}uided \textbf{E}xecution (SAGE), a two-stage framework that makes automated fine-tuning search cumulative (Figure~\ref{fig:intro}). In the first stage, SAGE conducts MCTS-based exploration with a role-specialized multi-agent pipeline, while a dedicated Distillation Agent transforms experimental outcomes into structured, reusable knowledge in an evolving experience repository. In the second stage, given a new task, SAGE retrieves experience that matches its requirements and applicability conditions, and uses the selected knowledge to guide the planning and execution of new fine-tuning experiments.

Each MCTS simulation here is an actual, hours-long training run. This cost rules out exhaustive search, so we adapt MCTS in several ways. Every scored node stays open to re-expansion instead of retiring once tried, and the backpropagated reward adds a parent-relative term so a genuine improvement is no longer scored the same as a regression that merely stays above baseline. On the knowledge side, the Distillation Agent extracts experience by comparing completed runs across the tree rather than any single result alone. This yields two tiers of experience: task-level exploration records that document each node's strategy and outcome, and
  confidence-scored cross-task insights that are updated deterministically and pruned when they fail validation.

Our core contributions are as follows:
\begin{itemize}
\item We propose SAGE, which accumulates structured experience from MCTS exploration and reuses it so a new task of the explored categories no longer needs to search from scratch.
\item We apply several adaptations to MCTS, including re-expandable nodes and the PaSR reward. These let it handle hours-long training runs and effectively discover strong strategies.
\item We design an experience mechanism that distills MCTS exploration into task-level records and confidence-scored cross-task insights, matched to new tasks through gap analysis and condition checks.
\item Across nine unseen tasks, SAGE's accumulated experience improves over baseline by an average of 12.4 percentage points more than the same pipeline without it, in a single round of execution.
\end{itemize}
  
\section{Related Work}

\paragraph{Automated Model Development.}
Automated model development aims to reduce human effort in model construction and training. Early work formalized algorithm selection and hyperparameter configuration as a joint optimization problem, solved with Bayesian optimization and evolutionary algorithms over predefined configuration spaces~\cite{thornton2013auto, feurer2015efficient, olson2016evaluation, falkner2018bohb}. Neural architecture search later extended this paradigm to the automatic design of deep model structures~\cite{DBLP:conf/iclr/ZophL17, real2019regularized, pham2018efficient, DBLP:conf/iclr/LiuSY19}. As large language models emerged, researchers began using semantic reasoning to guide algorithm configuration and experiment generation~\cite{DBLP:journals/corr/abs-2305-02499, DBLP:conf/iclr/LiuASS24}. LLM agents have since been applied to more open-ended machine learning engineering tasks, spanning tree-search- and MCTS-based pipeline exploration~\cite{DBLP:journals/corr/abs-2502-13138, DBLP:journals/corr/abs-2410-17238, DBLP:journals/corr/abs-2506-16499, toledo2026ai}, population-based evolutionary search~\cite{DBLP:journals/corr/abs-2512-24077, li2025fm}, and role-divided multi-agent automation~\cite{yang2025r, trirat2025automl, fang2026mlzero}. TREX~\cite{DBLP:journals/corr/abs-2604-14116} brings this paradigm to LLM fine-tuning, achieving competitive results across multiple tasks. However, TREX carries no experience across searches, so each new task must be solved from scratch. We further apply an experience mechanism to automated fine-tuning, proposing SAGE to accumulate structured cross-task experience through MCTS exploration and reuse it directly on new tasks.

\paragraph{Experiential Learning for Agents.}
Stateless agent designs ensure that experience accumulated during one task offers no benefit to subsequent tasks, motivating research into persistent experiential learning. Early work focused on within-task experience, where agents generate natural-language reflections across repeated attempts to guide subsequent executions~\cite{shinn2023reflexion, madaan2023self}, but such improvements do not transfer across tasks. Subsequent work extended the scope to cross-task settings, enabling agents to accumulate trajectories, abstract insights, or executable skills and retrieve them for new tasks via similarity-based lookup~\cite{DBLP:journals/tmlr/WangX0MXZFA24, zhao2024expel, DBLP:journals/pami/WangCLJHZLHZYML25}. This idea has been adapted to data science automation, where agents reuse historical experience through case retrieval, persistent knowledge bases, or hierarchical skill accumulation~\cite{guo2024ds, zhang2024mlcopilot, kim2026solve, DBLP:journals/corr/abs-2602-08234}.
SAGE extends this paradigm to automated LLM fine-tuning search. It structures experience into task-level exploration records and confidence-scored cross-task insights tailored to the fine-tuning strategy space.

\section{Method}

\begin{figure*}[t]
\centering
\includegraphics[width=\textwidth]{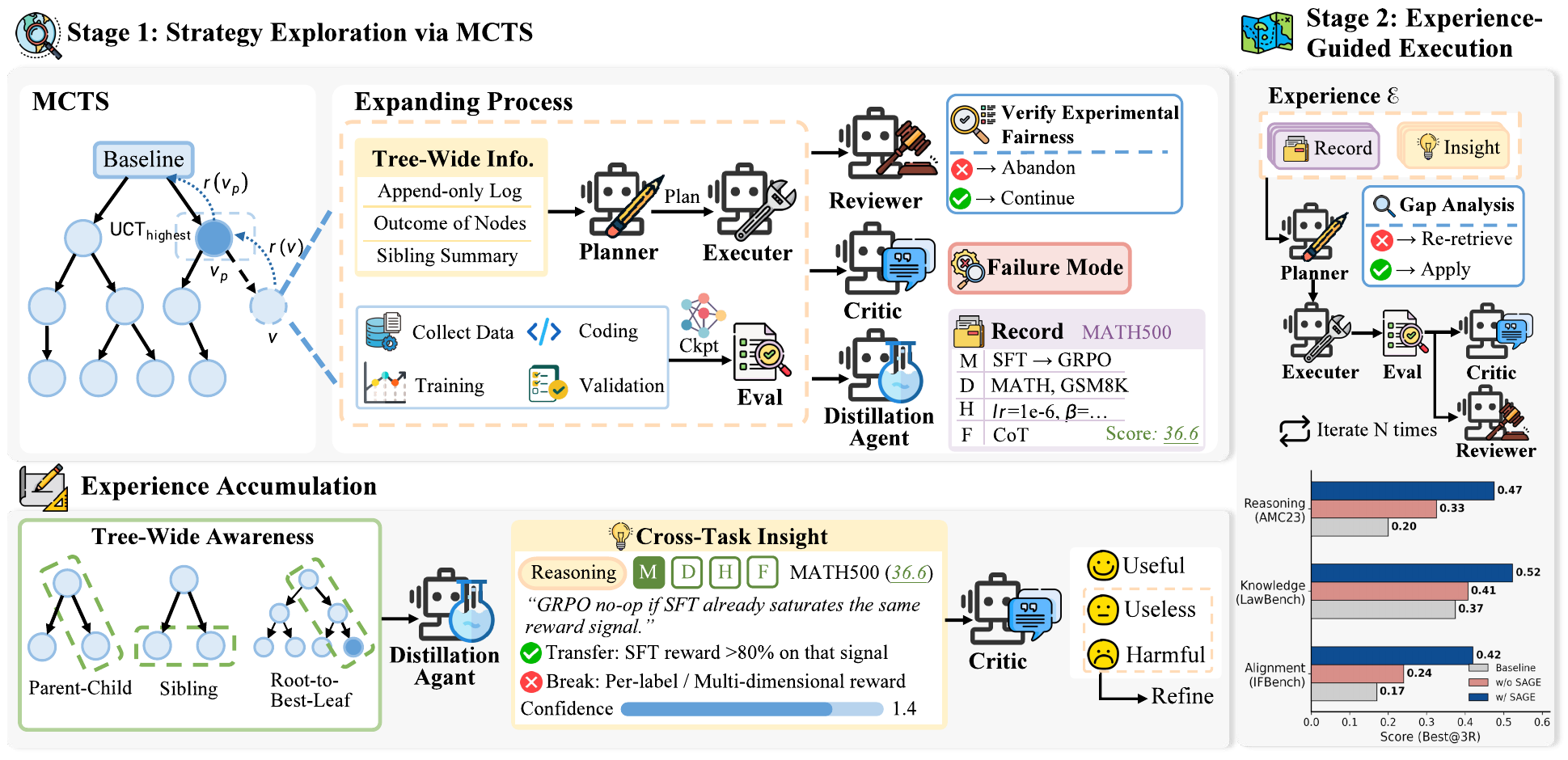}
\caption{Overview of SAGE. Stage~1 explores fine-tuning strategies via MCTS and distills them into structured experience; Stage~2 retrieves and applies this experience to guide execution on new tasks. Bars show representative Best@3R results per category (Table~\ref{tab:main_results}).}
\label{fig:overview}
\end{figure*}

In this section, we describe SAGE's design in detail. Figure~\ref{fig:overview} gives an overview of the full pipeline. We first formulate the problem and its strategy space (Section~\ref{sec:formulation}), then present MCTS-based strategy exploration (Section~\ref{sec:mcts}), two-tier structured experience accumulation (Section~\ref{sec:experience}), and experience-guided execution for new tasks (Section~\ref{sec:stage2}).

\subsection{Problem Formulation}
\label{sec:formulation}

Given an initial model $\theta_0$ and an evaluation function $\mathrm{eval}_\tau(\cdot)$ for task $\tau$, within the strategy space $\mathcal{S}=\mathcal{M}\times\mathcal{D}\times\mathcal{F}\times\mathcal{H}$, one fine-tuning decision reduces to solving:
\begin{equation}
\label{eq:objective}
\sigma^\star=\arg\max_{\sigma\in\mathcal{S}}\ \mathrm{eval}_\tau\big(\mathrm{train}(\theta_0,\sigma)\big),
\end{equation}
where $\sigma=(M,D,F,H)$. Here, $M$ denotes the training algorithm and phase composition (e.g., SFT, GRPO, or SFT-then-GRPO); $D$ denotes the source and scale of the training data; $F$ denotes how the data is constructed (e.g., CoT structure, system-prompt alignment); and $H$ denotes the hyperparameter configuration space. The four dimensions are coupled: the choice of $M$ constrains the range of $H$, and the distribution of $D$ shapes the design of $F$. This coupling rules out per-dimension search, motivating the MCTS-based exploration in Section~\ref{sec:mcts}.

We assume that fine-tuning strategies are more likely to transfer between tasks that share the same dominant capability bottleneck. In practice, we focus on three categories, $\mathcal{C}=\{\text{reasoning}, \text{knowledge}, \text{alignment}\}$, based on the capability bottleneck and its associated training paradigm. Reasoning tasks require learning reliable inference or solution-construction procedures, typically addressed through chain-of-thought data and reinforcement learning with verifiable rewards. Knowledge tasks require acquiring domain-specific facts and representations, typically through supervised fine-tuning on domain corpora. Alignment tasks require satisfying explicit instructions or behavioral constraints, typically through constraint-aware data construction and preference-based training. We denote by $g(\cdot)$ the function mapping a task to its associated categories, $g(\tau)\subseteq\mathcal{C}$, a non-empty subset; some tasks may span more than one category. Concrete task assignments appear in Section~\ref{sec:experiments}.

Stage 1 solves Eq.~\eqref{eq:objective} independently for a task $\tau$, and the search process is also distilled by the Distillation Agent into an experience update $\mathcal{E}\leftarrow\mathcal{E}\cup\Delta\mathcal{E}(\tau)$. Stage 2, facing a new task $\tau'$, no longer resolves the $\arg\max$ from scratch; instead, it retrieves category-matched exploration records and the full set of cross-task insights from $\mathcal{E}$ to directly plan and execute a strategy for $\tau'$. This replaces the full tree search of Stage 1 with experience-guided execution, reducing the search overhead required to reach a high-quality strategy.

\subsection{Strategy Exploration via MCTS}
\label{sec:mcts}
We formulate strategy search as Monte Carlo Tree Search (MCTS) over the strategy space $\mathcal{S}$. Each non-root node corresponds to one complete fine-tuning experiment: planned, trained, scored, and reviewed before it can grow further. Because each expansion is far costlier than the cheap rollouts classical MCTS assumes, we introduce five adaptations: an unbounded action space, tree-wide awareness, a diverge-refine trade-off, a multi-agent pipeline, and a Parent-Adjusted Scale-Referenced (PaSR) reward.

\paragraph{Unbounded Action Space.}
Classical MCTS assumes a finite action set and retires a node once all its actions have been tried. Fine-tuning strategies form an unbounded action space, so we generate actions on demand with a Planner agent: any scored node remains a candidate for further expansion. Selection therefore becomes a single global comparison rather than a level-by-level descent. We set $Q(v)=N(v)=0$ at node creation. While no reviewer-approved expansion has yet been recorded ($n=0$), selection uses the root directly. Each accepted child is backpropagated before any later UCT-based selection, so UCT is evaluated only for candidates with $N(v)>0$. Every scored node whose active child count is below a cap $k_{\max}$ is ranked by
\begin{equation}
\mathrm{UCT}(v)=
\begin{cases}
\dfrac{Q(v)}{N(v)}, & v=v_0,\\[4pt]
\dfrac{Q(v)}{N(v)}+c\sqrt{\dfrac{\ln N(v_{\mathrm{parent}})}{N(v)}}, & v\neq v_0,
\end{cases}
\label{eq:uct}
\end{equation}
and the highest-scoring node anywhere in the tree is chosen next. The root uses no exploration term because it has no parent. $Q(v)$ is the cumulative reward backpropagated through $v$, and $N(v)$ is its visit count; we define the reward in the PaSR Reward below.

\paragraph{Tree-Wide Awareness.}
An unbounded, agent-generated action space raises an obvious risk: nothing structurally prevents the Planner from re-proposing an already tried strategy. We close this gap with two layers of context passed to every Planner call. An append-only log records the direction and outcome of every completed or abandoned node in the tree, giving the Planner full search history before it proposes. A sibling summary further lists the direction and qualitative assessment of each existing child under the expanded node, for a tighter local comparison. This narrative context lets the Planner reason about precedent and avoid redundant strategies in a way a scalar visit count cannot.
  
\paragraph{Diverge-Refine Trade-off.}
UCT selects which node to expand next, but it does not specify how much the new child should depart from the selected node's strategy. We supply this signal with a continuous weight that combines two factors: how good the selected node already is, and how deep it sits in the tree. For the selected node $v$,
\begin{equation}
w(v) = \mathrm{clip}\big(1 - q(v) - d(v),\ 0,\ 1\big),
\label{eq:weight}
\end{equation}
where $q(v) = \frac{s(v) - s_{\text{base}}}{s_{\max} - s_{\text{base}}}$, clipped to $[0,1]$, measures $v$'s quality relative to the highest score $s_{\max}$ in the tree so far. When $s_{\max} \le s_{\text{base}}$, we default to favoring divergence rather than applying this normalization. $d(v) = \min(0.1 \cdot \mathrm{depth}(v),\, 0.3)$ adds a depth penalty that further biases deep nodes toward refinement, since they have already undergone multiple edits along a promising path. Low $w(v)$ instructs the Planner to refine $v$'s strategy, high $w(v)$ instructs it to diverge substantially, and intermediate values leave the choice to the Planner's discretion.

\paragraph{Multi-Agent Pipeline.}
Classical MCTS estimates a node's value with a cheap, disposable rollout; ours comes from an actual fine-tuning run, which demands substantial time and compute across distinct stages.   We therefore decompose this single step into a pipeline of role-isolated agents: a Planner proposes the strategy, an Executor implements it, a Reviewer verifies experimental fairness, and a Critic diagnoses failure modes to guide subsequent search. Deterministic evaluation scores the result between execution and review. We enforce role isolation with a sandbox: each agent's mount is read-write only for its designated outputs, everything else read-only. This prevents any agent from accidentally modifying other agents' artifacts or accessing data outside its permitted scope. A background Distillation stage then extracts reusable experience from the completed node, without blocking expansion elsewhere in the tree.

\paragraph{PaSR Reward.}
Classical MCTS backpropagates raw rollout outcomes. A baseline-relative reward is a natural alternative, but it conflates a genuine improvement over the parent with a regression that merely stays above baseline. We add a parent-relative term so the tree can distinguish the two:
\begin{equation}
r(v) = \Big[\underbrace{(s(v) - s_{\text{base}})}_{\text{baseline delta}} + \alpha \cdot \underbrace{(s(v) - s(v_{\text{parent}}))}_{\text{parent delta}}\Big] \cdot e(v),
\label{eq:pasr}
\end{equation}
where $e(v)$ is a GPU efficiency multiplier that discourages wasteful resource utilization, derived from memory occupancy and device coverage and lower-bounded at $0.5$ so that it down-weights but never zeroes out a positive reward; for a non-positive bracketed reward, we set $e(v)=1$. The baseline delta anchors every node to the same absolute reference, while the parent delta rewards incremental progress and penalizes regression along each search path.

\subsection{Experience Accumulation}
\label{sec:experience}
Without accumulated experience, the system exhibits two recurring failure modes. First, cold start: the Planner has no verified prior knowledge to draw on, so its early proposals remain inconsistent. Second, blind optimization: the Planner cannot tell which kinds of changes tend to help or hurt, so many proposed modifications regress rather than improve. Both trace back to a system that lacks transferable knowledge. We address this with a Distillation Agent that runs in the background, accumulating real execution outcomes as structured experience across two tiers.  Exploration records document the concrete strategy and outcome of every node in the tree. Cross-task insights distill transferable optimization patterns by contrasting different nodes, each refined under a confidence lifecycle.

\paragraph{Tree-Aware Distillation.}
A single completed node contributes its strategy and outcome to the exploration record, but is too isolated to attribute cause. Cross-task insights require contrast: what changed between two related runs, and what difference it made. The tree structure naturally supports such contrasts at multiple granularities. We exploit this through several Distillation Agent passes. A parent-child pass approximates a quasi-controlled comparison, since one edit typically touches only one or two strategy dimensions. A sibling pass compares divergent strategies from the same parent, revealing which dimension matters most. After the tree is complete, a root-to-best-leaf pass traces how the winning strategy evolved, and a whole-tree pass reviews all accumulated records and insights for completeness and correctness.

\paragraph{Exploration Records.}
Without prior experience, a new task lacks effective strategy guidance, leaving the Planner to propose blindly. Exploration records address this: as each node completes, the Distillation Agent records its strategy across method, dataset, format, and hyperparameters, together with its score. Records are organized per task but retrieved by category, so a new task reads the record from explored tasks in the same category, gaining a verified account of what worked and failed.

\begin{table*}[t]
              \setlength{\tabcolsep}{3pt}
              \begin{tabular}{@{}lllccccc@{}}
              \toprule
              \multirow{2}{*}{Category} & \multirow{2}{*}{Task} & \multirow{2}{*}{Metric} &
  \multirow{2}{*}{\shortstack{Baseline\\(Qwen2.5)}} & \multicolumn{2}{c}{w/o SAGE} &
    \multicolumn{2}{c}{w/ SAGE} \\
              \cmidrule(lr){5-6}\cmidrule(lr){7-8}
              & & & & Avg@3 & Best@3R & Avg@3 & Best@3R \\
              \midrule
              \multirow{2}{*}{Reasoning} & AMC23 & Acc. & 0.200 & 0.250\textsubscript{\makebox[4em][l]{\scriptsize+6.3\%}} &
    0.325\textsubscript{\makebox[4em][l]{\scriptsize+15.6\%}} & 0.425\textsubscript{\makebox[4em][l]{\scriptsize+28.1\%}} &
    0.475\textsubscript{\makebox[4em][l]{\scriptsize+34.4\%}} \\
               & HumanEval & Pass@1 & 0.122 & 0.234\textsubscript{\makebox[4em][l]{\scriptsize+12.8\%}} &
  0.354\textsubscript{\makebox[4em][l]{\scriptsize+26.4\%}} &
    0.354\textsubscript{\makebox[4em][l]{\scriptsize+26.4\%}} & 0.415\textsubscript{\makebox[4em][l]{\scriptsize+33.4\%}} \\
              \midrule
              \multirow{3}{*}{Knowledge} & LawBench & Avg. Score & 0.374 & 0.336\textsubscript{\makebox[4em][l]{\scriptsize-6.1\%}} &
    0.407\textsubscript{\makebox[4em][l]{\scriptsize+5.3\%}} & 0.399\textsubscript{\makebox[4em][l]{\scriptsize+4.0\%}} &
      0.523\textsubscript{\makebox[4em][l]{\scriptsize+23.8\%}} \\
               & ToMG-Bench & Val. \& Acc. & 0.211 & 0.362\textsubscript{\makebox[4em][l]{\scriptsize+19.1\%}} &
  0.491\textsubscript{\makebox[4em][l]{\scriptsize+35.5\%}}
    & 0.403\textsubscript{\makebox[4em][l]{\scriptsize+24.3\%}} & 0.660\textsubscript{\makebox[4em][l]{\scriptsize+56.9\%}} \\
               & OpenFindata & Acc. & 0.602 & 0.557\textsubscript{\makebox[4em][l]{\scriptsize-11.3\%}} &
  0.585\textsubscript{\makebox[4em][l]{\scriptsize-4.3\%}} &
    0.639\textsubscript{\makebox[4em][l]{\scriptsize+9.3\%}} & 0.651\textsubscript{\makebox[4em][l]{\scriptsize+12.3\%}} \\
              \midrule
             Alignment & IFBench & Acc.$_{\text{prompt-loose}}$ & 0.170 & 0.164\textsubscript{\makebox[4em][l]{\scriptsize-0.7\%}} &
    0.240\textsubscript{\makebox[4em][l]{\scriptsize+8.4\%}} & 0.244\textsubscript{\makebox[4em][l]{\scriptsize+8.9\%}} &
    0.420\textsubscript{\makebox[4em][l]{\scriptsize+30.1\%}} \\
              \midrule
              \multirow{2}{*}{Reasoning \& Knowledge} & GPQA-Diamond & Acc. & 0.131 &
  0.170\textsubscript{\makebox[4em][l]{\scriptsize+4.5\%}} &
    0.253\textsubscript{\makebox[4em][l]{\scriptsize+14.0\%}} & 0.259\textsubscript{\makebox[4em][l]{\scriptsize+14.7\%}} &
      0.308\textsubscript{\makebox[4em][l]{\scriptsize+20.4\%}} \\
               & EconLogicQA & Acc. & 0.039 & 0.049\textsubscript{\makebox[4em][l]{\scriptsize+1.0\%}} &
  0.108\textsubscript{\makebox[4em][l]{\scriptsize+7.2\%}} &
    0.151\textsubscript{\makebox[4em][l]{\scriptsize+11.7\%}} & 0.177\textsubscript{\makebox[4em][l]{\scriptsize+14.4\%}} \\
              \midrule
              Knowledge \& Alignment & ACI-Bench & Rouge-1 & 0.191 & 0.215\textsubscript{\makebox[4em][l]{\scriptsize+3.0\%}} &
    0.362\textsubscript{\makebox[4em][l]{\scriptsize+21.1\%}} & 0.294\textsubscript{\makebox[4em][l]{\scriptsize+12.7\%}} &
    0.493\textsubscript{\makebox[4em][l]{\scriptsize+37.3\%}}
      \\
              \bottomrule
              \end{tabular}
              \caption{Main results on nine Stage~2 tasks. Baseline is the initial model $\theta_0$ before fine-tuning. w/o SAGE runs the
  multi-agent
    pipeline without experience; w/ SAGE adds full experience retrieval. Subscripts show relative improvement
  $\Delta(s)$
    (Eq.~\ref{eq:relimp}).}
          \label{tab:main_results}
              \end{table*}

\paragraph{Cross-Task Insights.}
Exploration records are concrete but task-specific, leaving it to the consuming agent to judge whether an outcome transfers. Cross-task insights instead distill a recurring empirical pattern along the same four dimensions together with the conditions under which the pattern holds or breaks. Stating these conditions explicitly is what makes an insight safe to reuse: a later task can verify them directly rather than guessing whether a pattern learned elsewhere still applies.

Each insight carries a confidence score that evolves as evidence accumulates. Whenever an insight is referenced in a completed experiment, the Critic labels the outcome as useful, useless, or harmful. Let $v_i$ denote the $i$-th verdict; confidence is recomputed on every update:
\[
c_0 = c_{\text{init}}, \qquad
c_i =
\begin{cases}
c_{i-1} + \delta & \text{if } v_i = \text{useful}, \\
c_{i-1} - \delta & \text{if } v_i = \text{useless}, \\
\gamma \cdot c_{i-1} & \text{if } v_i = \text{harmful},
\end{cases}
\]
where $\gamma < 1$ makes the penalty for a harmful outcome multiplicative, so a single contradicting experiment can outweigh several confirmations. A useless or harmful verdict can also prompt the Distillation Agent to revise the insight's content, though the confidence effect is determined solely by the verdict sequence above. These verdicts accumulate during MCTS search itself, so confidence is already updating before the search ends. Any insight whose confidence falls below a fixed threshold is pruned automatically.

\subsection{Experience-Guided Execution}
\label{sec:stage2}
Stage~1 explores each meta-task through branching search, trying many strategies so the Distillation Agent can compare them and extract what actually helps. Stage~2 is designed so a new task reaches a strong result in as few rounds as possible, guided by the experience Stage~1 has accumulated rather than by a search of its own. We therefore reuse the same pipeline (Planner, Executor, Eval, Reviewer, Critic) in a short, fixed sequence of rounds. After Round~1, rounds carry only the previous plan and score-free Critic report; each Executor run applies the revised plan afresh to $\theta_0$, carrying neither fine-tuned weights nor exact scores.

Each round's Planner begins by retrieving experience for every category in the task's type. Because Stage~1 explores one meta-task per category, this yields one exploration record per matched category plus the full set of cross-task insights. However, a shared category label does not guarantee that the same capability is being tested, so before adopting any experience the Planner runs a gap analysis: it compares the current task's baseline, evaluation mechanism, and headroom against those of each source task to judge whether the source's strategy scale applies here. Adopting a cross-task insight carries a further check, verifying the conditions the insight itself states, since a pattern that holds for one task can break for another even within the same category.

\section{Experiments}
\label{sec:experiments}

This section evaluates whether SAGE's accumulated experience improves automated fine-tuning on unseen tasks. We first describe the experimental setup, including tasks, models, and evaluation protocols (Section~\ref{sec:setup}). We then compare the full pipeline with and without SAGE's experience mechanism across nine Stage~2 tasks (Section~\ref{sec:main_results}).
  
\subsection{Experimental Setup}
\label{sec:setup}

\paragraph{Tasks.}
Using the three categories from Section~\ref{sec:formulation}: reasoning, knowledge, and alignment, we split the evaluation into two stages.
Stage~1 selects one meta-task per category for MCTS exploration and experience accumulation: MATH500~\cite{hendrycks2021measuring, lightman2024let} (reasoning), HoC~\cite{DBLP:journals/bioinformatics/BakerSGAHSK16} (knowledge), and IFEval~\cite{DBLP:journals/corr/abs-2311-07911} (alignment). Stage~2 evaluates on nine unseen tasks spanning both single- and cross-category settings: AMC23 and HumanEval~\cite{chen2021evaluating}; LawBench~\cite{fei2024lawbench}, ToMG-Bench~\cite{DBLP:journals/corr/abs-2412-14642}, and OpenFindata~\cite{openfindata2023}; IFBench~\cite{DBLP:journals/corr/abs-2507-02833}; GPQA-Diamond~\cite{DBLP:journals/corr/abs-2311-12022} and EconLogicQA~\cite{quan2024econlogicqa}; and ACI-Bench~\cite{DBLP:journals/corr/abs-2306-02022}.

\paragraph{System setting.}
We use GLM-5.1~\cite{zeng2026glm} to drive every agent in the pipeline. For the initial model $\theta_0$, tasks whose category includes reasoning use Qwen2.5-Math-1.5B-Instruct~\cite{DBLP:journals/corr/abs-2409-12122}, and all other tasks use Qwen2.5-1.5B-Instruct~\cite{DBLP:journals/corr/abs-2412-15115}. All training runs use ms-swift~\cite{zhao2024swiftascalablelightweightinfrastructure} as the fine-tuning framework.

\paragraph{Evaluation protocol.}
We report results under two complementary protocols. Avg@3 averages three independent Round~1 runs and measures stability under the randomness inherent in automated fine-tuning. An initial Round~1 run is continued through Rounds~2--3 while the other two Round~1 runs are executed in parallel, so the refinement chain is fixed before their outcomes are observed rather than selected by score. Best@3R takes the maximum score from one three-round refinement chain, which carries the previous plan and score-free Critic report but restarts every Executor run from $\theta_0$; task-specific metrics are listed in Table~\ref{tab:main_results}. Each score in the table is further annotated with its relative improvement over baseline:
\begin{equation}
\Delta(s) = \frac{s - s_{\text{base}}}{1 - s_{\text{base}}},
\label{eq:relimp}
\end{equation}
which normalizes the gain by the remaining headroom to a perfect score.

\subsection{Main Results}
\label{sec:main_results}
Table~\ref{tab:main_results} presents results for our full pipeline with and without SAGE's experience mechanism across the nine Stage~2 tasks, together with each score's improvement relative to the initial model $\theta_0$.

\paragraph{Consistent gains across all tasks.} Across all nine tasks, w/ SAGE outperforms w/o SAGE under both Avg@3 and Best@3R. On average, w/ SAGE improves over baseline by 15.6\% under Avg@3 and 29.2\% under Best@3R, compared to 3.2\% and 14.4\% for w/o SAGE---a 12.4-point gain in single-round execution that widens to 14.8 points after a short refinement sequence. The gap is largest on AMC23 under Avg@3, where w/ SAGE improves over baseline by 28.1\%, versus only 6.3\% for w/o SAGE. It is also largest on IFBench under Best@3R, where w/ SAGE reaches 30.1\%, versus only 8.4\% for w/o SAGE. This consistency across nine tasks suggests that SAGE's accumulated experience provides a robust source of guidance beyond incidental task-to-task variation.

\paragraph{Risk of blind optimization without experience.} Without accumulated experience, the multi-agent pipeline can regress below the baseline it starts from. Under Avg@3, w/o SAGE falls 6.1\% below baseline on LawBench. On OpenFindata the drop is larger, 11.3\% below baseline. This matches the blind-optimization risk described in Section~\ref{sec:experience}: without prior guidance, proposed changes are essentially blind, and some even make the model worse. On both tasks, w/ SAGE stays above baseline under the same protocol, suggesting that accumulated experience can help avoid such regressions.

\paragraph{Single- vs. cross-category tasks.} GPQA-Diamond, EconLogicQA, and ACI-Bench each pair two base categories, so guiding them requires combining experience from two sources rather than reusing one directly. On these three tasks, the average gain from adding SAGE is 10.2\% under Avg@3 and 9.9\% under Best@3R, smaller than the 13.5\% and 17.3\% averaged over the six single-category tasks. The largest cross-category gain is on ACI-Bench under Best@3R, where w/ SAGE improves over baseline by 37.3\%, versus 21.1\% for w/o SAGE. SAGE still improves every cross-category task under both protocols, but the smaller margin suggests that combining experience across categories, where condition mismatches are more likely, yields less direct guidance than reusing it within one.

\section{Analysis}
\subsection{Strategy Accumulation Analysis}
\label{sec:stage1_analysis}

Having evaluated accumulated experience on downstream tasks, we now examine Stage~1 itself: whether MCTS search produces experience worth accumulating, and whether the confidence mechanism in Section~\ref{sec:experience} really works.

\begin{wraptable}{r}{0.50\textwidth}
\centering
\setlength{\columnwidth}{\linewidth}
\captionsetup{width=\linewidth}
\small
\begin{tabular}{@{}llcc@{}}
\toprule
Meta-task & Category & Baseline & Best \\
\midrule
MATH500 & Reasoning & 0.366 & 0.732\textsubscript{\makebox[2.5em][l]{\scriptsize+57.7\%}} \\
HoC & Knowledge & 0.294 & 0.842\textsubscript{\makebox[2.5em][l]{\scriptsize+77.6\%}} \\
IFEval & Alignment & 0.388 & 0.621\textsubscript{\makebox[2.5em][l]{\scriptsize+38.1\%}} \\
\bottomrule
\end{tabular}
\caption{MCTS search summary on the three Stage~1 meta-tasks. Subscripts denote relative improvement over baseline, as in Table~\ref{tab:main_results}.}
\label{tab:mcts_search}
\end{wraptable}

\begin{wrapfigure}[14]{r}{0.48\textwidth}
\centering
\setlength{\columnwidth}{\linewidth}
\captionsetup{width=\linewidth}
\includegraphics[width=\linewidth]{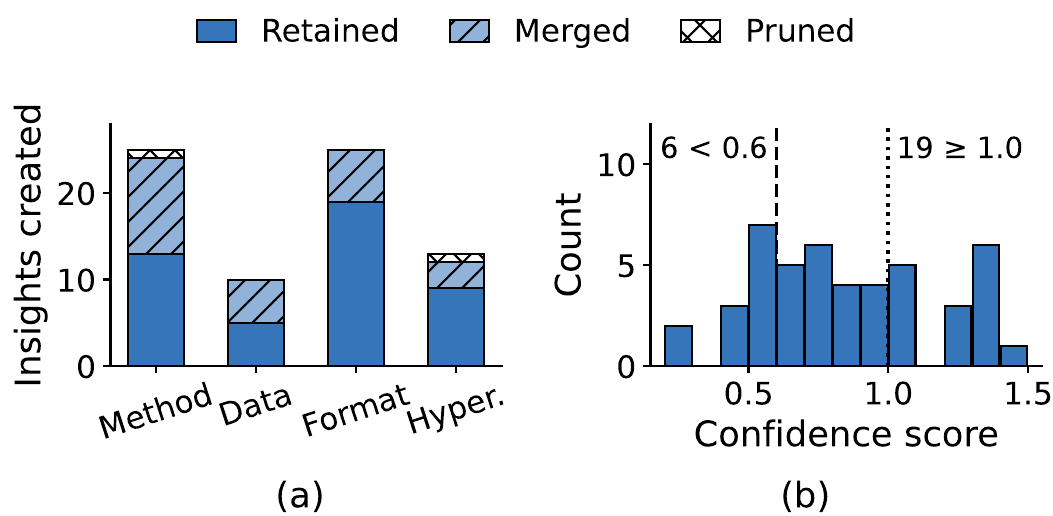}
\caption{Cross-task insight lifecycle. (a) Of 73 insights, 46 were retained, 25 merged, and 2 pruned. (b) Final confidence; labels count insights validated at least four times.}
\label{fig:insight_lifecycle}
\end{wrapfigure}

\paragraph{Stage 1 overview.} Each meta-task runs 20 MCTS expansions, substantially improving over baseline on all three Stage~1 tasks (Table~\ref{tab:mcts_search}). The Distillation Agent turns this search into the two-tier experience structure from Section~\ref{sec:experience}: three per-task exploration records and a final set of 46 cross-task insights. Of the 73 insights ever created, 27 did not survive: 25 were merged into a related insight as near-duplicates, and 2 were pruned automatically for falling below the confidence threshold. Figure~\ref{fig:insight_lifecycle}(a) breaks this down by dimension: Format insights survive at the highest rate (19/25, 76\%), while Method and Dataset insights are merged away roughly as often as they are kept.

\Needspace{0.34\textheight}

\paragraph{Continually self-correcting insights.} Confidence on these insights does not simply rise with more evidence. Across 228 validation verdicts, 74 (32.5\%) revise, weaken, or contradict the existing insight instead of confirming it. Six insights stay below a confidence of 0.6 even after at least four rounds of validation. Nineteen insights (41.3\%) sit at the opposite extreme: a confidence at or above 1.0, ranging up to 1.5, each likewise requiring at least four validation rounds to reach. Figure~\ref{fig:insight_lifecycle}(b) shows this distribution directly. This shows the confidence lifecycle in Section~\ref{sec:experience} performs genuine quality control rather than uncritical accumulation.

\subsection{Guidance from Critic and Experience}
\label{sec:ablation_critic}

Beyond accumulated experience, the Critic offers a second kind of guidance within SAGE. Whereas accumulated experience transfers knowledge across tasks, the Critic draws on the current task's most recent round to guide subsequent refinement (Section~\ref{sec:mcts}). This section builds the pipeline up in three stages---neither source active (w/o Critic), Critic only (w/o SAGE), and both (w/ SAGE)---and checks what each adds to Table~\ref{tab:main_results}.

\begin{wrapfigure}{r}{0.48\textwidth}
\centering
\setlength{\columnwidth}{\linewidth}
\captionsetup{width=\linewidth}
\includegraphics[width=\linewidth]{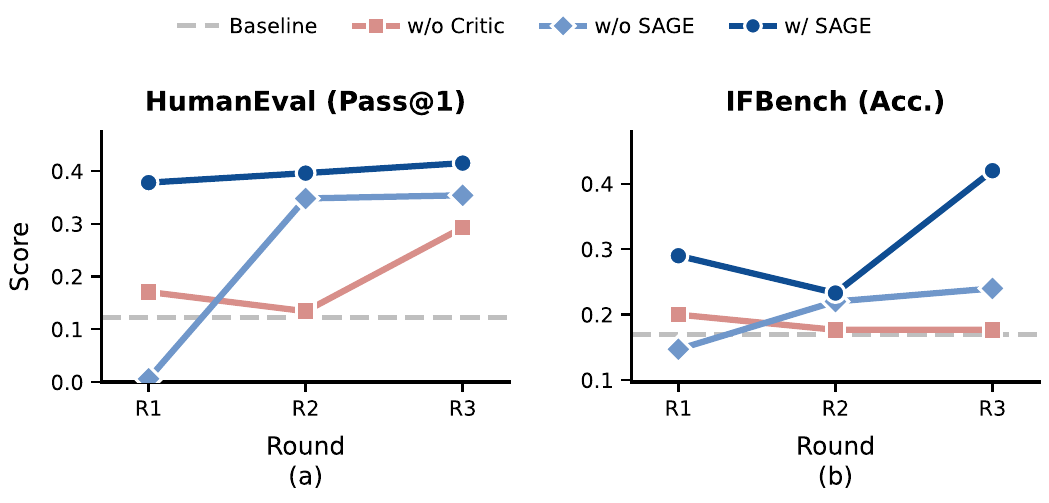}
\caption{Round-wise scores for w/o Critic, w/o SAGE, and w/ SAGE under the three-round refinement protocol. (a) HumanEval. (b) IFBench.}
\label{fig:ablation_critic}
\end{wrapfigure}

\paragraph{Setup and results.} We report HumanEval and IFBench as illustrative cases under the same Best@3R protocol as Table~\ref{tab:main_results} through all three configurations. Figure~\ref{fig:ablation_critic} plots each configuration's score across all three rounds. On both tasks, w/ SAGE reaches the highest Round~3 score (0.415 on HumanEval, 0.420 on IFBench), followed by w/o SAGE (0.354, 0.240) and w/o Critic (0.293, 0.177).

\paragraph{Distinct roles for each signal.} The Critic and experience improve results differently: the Critic keeps refinement from failing round to round, while experience changes a specific round's own performance. Without the Critic, refinement stalls: on IFBench its best score remains the Round~1 value of 0.200, and on HumanEval it recovers only partially. With the Critic restored, w/o SAGE improves every round on both tasks. Adding experience in turn raises Round 1's own score: w/ SAGE reaches 0.378 on HumanEval and 0.290 on IFBench at Round 1 alone, already beating w/o SAGE's full three-round best (0.354, 0.240). The Critic helps optimization improve more stably, while experience further accelerates the discovery of an effective strategy.

\Needspace{0.34\textheight}
\subsection{GPU Efficiency}
\label{sec:gpu_eff}

\begin{wraptable}{r}{0.50\textwidth}
\centering
\setlength{\columnwidth}{\linewidth}
\captionsetup{width=\linewidth}
\small
\begin{tabular}{@{}lccc@{}}
\toprule
Protocol & w/o SAGE & w/ SAGE & $\Delta$ \\
\midrule
Avg@3  & 0.213 & 0.338 & +0.125 \\
Avg@3R & 0.233 & 0.336 & +0.103 \\
\bottomrule
\end{tabular}
\caption{Average GPU memory utilization on Stage~2 tasks. Avg@3 matches Table~\ref{tab:main_results}. Avg@3R averages, rather than maxes, over the same three rounds used by Best@3R.}
\label{tab:gpu_eff}
\end{wraptable}

We further examine whether the resource-utilization practices accumulated during Stage~1 transfer to Stage~2. We manually inspected the accumulated experience and found that the MCTS reward encourages the discovery of strategies that improve GPU resource utilization. One representative example is disabling the default gradient checkpointing when memory permits, which increases GPU memory occupancy from 55\% to 92\%. Such efficiency-oriented experience is retained in the repository and subsequently retrieved by the Stage 2 planner when its applicability conditions are satisfied, contributing to the higher GPU memory utilization. As shown in Table~\ref{tab:gpu_eff}, w/~SAGE achieves consistently higher utilization under both protocols.

\subsection{Case Study}
\label{sec:case_study}

We trace method selection on IFBench (baseline 0.170) to illustrate how experience shapes strategy.
Both conditions select SFT$\to$GRPO, but experience produces a fundamentally different instantiation. Guided by an insight that unverified data poisons constraint learning, the w/~SAGE planner verifies all training data programmatically. It also builds a custom GRPO reward from IFBench's own constraint-verification library, since gap analysis shows that IFBench's lower baseline (0.170 vs.\ IFEval's 0.388) leaves room for GRPO to contribute. Without experience, the planner selects instruction-following datasets whose constraint types have zero overlap with IFBench's 58 novel test types. Scores: 0.290 vs.\ 0.147.

In subsequent rounds, the Critic in both conditions flags the shortcomings of Round~1, but the two planners respond in opposite directions. Drawing on experience, the w/~SAGE planner judges the training method sound and attributes the gap to insufficient coverage of rare constraint types. It augments training data accordingly and reaches 0.420 by Round~3. Without experience, the planner attributes the gap to method choice instead, cycling from GRPO to DPO and back without addressing the underlying data mismatch, reaching only 0.240.

\subsection{Out-of-Category Generalization}
\label{sec:ood}

\begin{wraptable}{R}{0.50\textwidth}
\centering
\setlength{\columnwidth}{\linewidth}
\captionsetup{width=\linewidth}
\small
\begin{tabular}{@{}llccc@{}}
\toprule
Task & Protocol & Baseline & w/o SAGE & w/ SAGE \\
\midrule
\multirow{2}{*}{WMT24PP} & Avg@3 & \multirow{2}{*}{31.49} & 27.30 & 30.79 \\
& Best@3R & & 28.96 & 35.06 \\
\midrule
\multirow{2}{*}{FLORES} & Avg@3 & \multirow{2}{*}{35.73} & 28.58 & 37.96 \\
 & Best@3R & & 35.07 & 40.64 \\
\bottomrule
\end{tabular}
\caption{Out-of-category generalization on English-to-Simplified-Chinese translation (SacreBLEU).}
\label{tab:ood}
\end{wraptable}

The nine Stage~2 tasks in Section~\ref{sec:main_results} all belong to categories explored in Stage~1. To test whether accumulated experience generalizes beyond these categories, we select two translation datasets---WMT24PP~\cite{deutsch2025wmt24++} and FLORES~\cite{costa2022no}---and evaluate on their English-to-Simplified-Chinese splits. Translation falls outside all three Stage~1 categories, so no exploration records exist; the Planner draws only on cross-task insights.

As shown in Table~\ref{tab:ood}, w/~SAGE surpasses baseline under Best@3R on both tasks and under Avg@3 on FLORES, whereas w/o~SAGE regresses on every metric. Nevertheless, cross-task insights alone still recover most of the gap created by the experience-free pipeline, suggesting that they can transfer useful fine-tuning knowledge beyond the categories from which they were derived.

\section{Conclusion}

We presented SAGE, a two-stage framework that makes automated LLM fine-tuning cumulative rather than restarting each task from scratch. In Stage~1, an adapted MCTS-based multi-agent pipeline explores fine-tuning strategies, while a Distillation Agent converts experimental outcomes into task-level exploration records and confidence-scored cross-task insights. In Stage~2, SAGE retrieves relevant experience and applies it through gap analysis and condition verification to guide new tasks. Across nine unseen tasks, SAGE increases the average relative improvement from 3.2\% to 15.6\% under Avg@3 and from 14.4\% to 29.2\% under Best@3R. These results demonstrate that persistent, structured experience can reduce repeated search and accelerate automated fine-tuning on unseen tasks.

\bibliography{main}

@article{li2023chatdoctor,
  title={Chatdoctor: A medical chat model fine-tuned on a large language model meta-ai (llama) using medical domain knowledge},
  author={Li, Yunxiang and Li, Zihan and Zhang, Kai and Dan, Ruilong and Jiang, Steve and Zhang, You},
  journal={Cureus},
  volume={15},
  number={6},
  year={2023},
  publisher={Cureus}
}

@article{roziere2023code,
  title={Code llama: Open foundation models for code},
  author={Roziere, Baptiste and Gehring, Jonas and Gloeckle, Fabian and Sootla, Sten and Gat, Itai and Tan, Xiaoqing Ellen and Adi, Yossi and Liu, Jingyu and Sauvestre, Romain and Remez, Tal and others},
  journal={arXiv preprint arXiv:2308.12950},
  year={2023}
}

@article{DBLP:journals/corr/abs-2502-13138,
  author       = {Zhengyao Jiang and
                  Dominik Schmidt and
                  Dhruv Srikanth and
                  Dixing Xu and
                  Ian Kaplan and
                  Deniss Jacenko and
                  Yuxiang Wu},
  title        = {{AIDE:} AI-Driven Exploration in the Space of Code},
  journal      = {CoRR},
  volume       = {abs/2502.13138},
  year         = {2025},
  url          = {https://doi.org/10.48550/arXiv.2502.13138},
  doi          = {10.48550/ARXIV.2502.13138},
  eprinttype   = {arXiv},
  eprint       = {2502.13138},
  bibsource    = {dblp computer science bibliography, https://dblp.org}
}

@article{DBLP:journals/corr/abs-2506-16499,
  author       = {Zexi Liu and
                  Yuzhu Cai and
                  Xinyu Zhu and
                  Yujie Zheng and
                  Runkun Chen and
                  Ying Wen and
                  Yanfeng Wang and
                  Weinan E and
                  Siheng Chen},
  title        = {ML-Master: Towards AI-for-AI via Integration of Exploration and Reasoning},
  journal      = {CoRR},
  volume       = {abs/2506.16499},
  year         = {2025},
  url          = {https://doi.org/10.48550/arXiv.2506.16499},
  doi          = {10.48550/ARXIV.2506.16499},
  eprinttype   = {arXiv},
  eprint       = {2506.16499},
  bibsource    = {dblp computer science bibliography, https://dblp.org}
}

@article{DBLP:journals/corr/abs-2604-13018,
  author       = {Guoxin Chen and
                  Jie Chen and
                  Lei Chen and
                  Jiale Zhao and
                  Fanzhe Meng and
                  Wayne Xin Zhao and
                  Ruihua Song and
                  Cheng Chen and
                  Ji{-}Rong Wen and
                  Kai Jia},
  title        = {Toward Autonomous Long-Horizon Engineering for {ML} Research},
  journal      = {CoRR},
  volume       = {abs/2604.13018},
  year         = {2026},
  url          = {https://doi.org/10.48550/arXiv.2604.13018},
  doi          = {10.48550/ARXIV.2604.13018},
  eprinttype   = {arXiv},
  eprint       = {2604.13018},
  bibsource    = {dblp computer science bibliography, https://dblp.org}
}

@article{DBLP:journals/corr/abs-2604-14116,
  author       = {Zerun Ma and
                  Guoyin Wang and
                  Xinchen Xie and
                  Yicheng Chen and
                  He Du and
                  Bowen Li and
                  Yanan Sun and
                  Wenran Liu and
                  Kai Chen and
                  Yining Li},
  title        = {{TREX:} Automating {LLM} Fine-tuning via Agent-Driven Tree-based Exploration},
  journal      = {CoRR},
  volume       = {abs/2604.14116},
  year         = {2026},
  url          = {https://doi.org/10.48550/arXiv.2604.14116},
  doi          = {10.48550/ARXIV.2604.14116},
  eprinttype   = {arXiv},
  eprint       = {2604.14116},
  bibsource    = {dblp computer science bibliography, https://dblp.org}
}

@article{feurer2015efficient,
  title={Efficient and robust automated machine learning},
  author={Feurer, Matthias and Klein, Aaron and Eggensperger, Katharina and Springenberg, Jost and Blum, Manuel and Hutter, Frank},
  journal={Advances in neural information processing systems},
  volume={28},
  year={2015}
}

@inproceedings{thornton2013auto,
  title={Auto-WEKA: Combined selection and hyperparameter optimization of classification algorithms},
  author={Thornton, Chris and Hutter, Frank and Hoos, Holger H and Leyton-Brown, Kevin},
  booktitle={Proceedings of the 19th ACM SIGKDD international conference on Knowledge discovery and data mining},
  pages={847--855},
  year={2013}
}

@inproceedings{olson2016evaluation,
  title={Evaluation of a tree-based pipeline optimization tool for automating data science},
  author={Olson, Randal S and Bartley, Nathan and Urbanowicz, Ryan J and Moore, Jason H},
  booktitle={Proceedings of the genetic and evolutionary computation conference 2016},
  pages={485--492},
  year={2016}
}

@inproceedings{falkner2018bohb,
  title={BOHB: Robust and efficient hyperparameter optimization at scale},
  author={Falkner, Stefan and Klein, Aaron and Hutter, Frank},
  booktitle={International conference on machine learning},
  pages={1437--1446},
  year={2018},
  organization={PMLR}
}

@inproceedings{DBLP:conf/iclr/ZophL17,
  author       = {Barret Zoph and
                  Quoc V. Le},
  title        = {Neural Architecture Search with Reinforcement Learning},
  booktitle    = {5th International Conference on Learning Representations, {ICLR} 2017,
                  Toulon, France, April 24-26, 2017, Conference Track Proceedings},
  publisher    = {OpenReview.net},
  year         = {2017},
  url          = {https://openreview.net/forum?id=r1Ue8Hcxg},
  bibsource    = {dblp computer science bibliography, https://dblp.org}
}

@inproceedings{real2019regularized,
  title={Regularized evolution for image classifier architecture search},
  author={Real, Esteban and Aggarwal, Alok and Huang, Yanping and Le, Quoc V},
  booktitle={Proceedings of the aaai conference on artificial intelligence},
  volume={33},
  number={01},
  pages={4780--4789},
  year={2019}
}

@inproceedings{pham2018efficient,
  title={Efficient neural architecture search via parameters sharing},
  author={Pham, Hieu and Guan, Melody and Zoph, Barret and Le, Quoc and Dean, Jeff},
  booktitle={International conference on machine learning},
  pages={4095--4104},
  year={2018},
  organization={PMLR}
}

@inproceedings{DBLP:conf/iclr/LiuSY19,
  author       = {Hanxiao Liu and
                  Karen Simonyan and
                  Yiming Yang},
  title        = {{DARTS:} Differentiable Architecture Search},
  booktitle    = {7th International Conference on Learning Representations, {ICLR} 2019,
                  New Orleans, LA, USA, May 6-9, 2019},
  publisher    = {OpenReview.net},
  year         = {2019},
  url          = {https://openreview.net/forum?id=S1eYHoC5FX},
  bibsource    = {dblp computer science bibliography, https://dblp.org}
}

@article{DBLP:journals/corr/abs-2305-02499,
  author       = {Shujian Zhang and
                  Chengyue Gong and
                  Lemeng Wu and
                  Xingchao Liu and
                  Mingyuan Zhou},
  title        = {AutoML-GPT: Automatic Machine Learning with {GPT}},
  journal      = {CoRR},
  volume       = {abs/2305.02499},
  year         = {2023},
  url          = {https://doi.org/10.48550/arXiv.2305.02499},
  doi          = {10.48550/ARXIV.2305.02499},
  eprinttype   = {arXiv},
  eprint       = {2305.02499},
  bibsource    = {dblp computer science bibliography, https://dblp.org}
}

@inproceedings{DBLP:conf/iclr/LiuASS24,
  author       = {Tennison Liu and
                  Nicol{\'{a}}s Astorga and
                  Nabeel Seedat and
                  Mihaela van der Schaar},
  title        = {Large Language Models to Enhance Bayesian Optimization},
  booktitle    = {The Twelfth International Conference on Learning Representations,
                  {ICLR} 2024, Vienna, Austria, May 7-11, 2024},
  publisher    = {OpenReview.net},
  year         = {2024},
  url          = {https://openreview.net/forum?id=OOxotBmGol},
  bibsource    = {dblp computer science bibliography, https://dblp.org}
}

@article{toledo2026ai,
  title={Ai research agents for machine learning: Search, exploration, and generalization in mle-bench},
  author={Toledo, Edan and Hambardzumyan, Karen and Josifoski, Martin and Hazra, Rishi and Baldwin, Nicolas and Audran-Reiss, Alexis and Kuchnik, Michael and Magka, Despoina and Jiang, Minqi and Lupidi, Alisia and others},
  journal={Advances in Neural Information Processing Systems},
  volume={38},
  pages={35309--35348},
  year={2026}
}

@article{DBLP:journals/corr/abs-2512-24077,
  author       = {Chunhui Wan and
                  Xunan Dai and
                  Zhuo Wang and
                  Minglei Li and
                  Yanpeng Wang and
                  Yinan Mao and
                  Yu Lan and
                  Zhiwen Xiao},
  title        = {LoongFlow: Directed Evolutionary Search via a Cognitive Plan-Execute-Summarize
                  Paradigm},
  journal      = {CoRR},
  volume       = {abs/2512.24077},
  year         = {2025},
  url          = {https://doi.org/10.48550/arXiv.2512.24077},
  doi          = {10.48550/ARXIV.2512.24077},
  eprinttype   = {arXiv},
  eprint       = {2512.24077},
  bibsource    = {dblp computer science bibliography, https://dblp.org}
}

@article{li2025fm,
  title={The fm agent},
  author={Li, Annan and Wu, Chufan and Ge, Zengle and Chong, Yee Hin and Hou, Zhinan and Cao, Lizhe and Ju, Cheng and Wu, Jianmin and Li, Huaiming and Zhang, Haobo and others},
  journal={arXiv preprint arXiv:2510.26144},
  year={2025}
}

@article{yang2025r,
  title={R\&D-Agent: An LLM-Agent Framework Towards Autonomous Data Science},
  author={Yang, Xu and Yang, Xiao and Fang, Shikai and Zhang, Yifei and Wang, Jian and Xian, Bowen and Li, Qizheng and Li, Jingyuan and Xu, Minrui and Li, Yuante and others},
  journal={arXiv preprint arXiv:2505.14738},
  year={2025}
}

@inproceedings{trirat2025automl,
  title={AutoML-Agent: A Multi-Agent LLM Framework for Full-Pipeline AutoML},
  author={Trirat, Patara and Jeong, Wonyong and Hwang, Sung Ju},
  booktitle={International Conference on Machine Learning},
  pages={60099--60146},
  year={2025},
  organization={PMLR}
}

@article{fang2026mlzero,
  title={Mlzero: A multi-agent system for end-to-end machine learning automation},
  author={Fang, Haoyang and Han, Boran and Erickson, Nick and Zhang, Xiyuan and Zhou, Su and Dagar, Anirudh and Zhang, Jiani and Turkmen, Ali Caner and Hu, Tony and Rangwala, Huzefa and others},
  journal={Advances in Neural Information Processing Systems},
  volume={38},
  pages={69001--69070},
  year={2026}
}

@article{DBLP:journals/corr/abs-2410-17238,
  author       = {Yizhou Chi and
                  Yizhang Lin and
                  Sirui Hong and
                  Duyi Pan and
                  Yaying Fei and
                  Guanghao Mei and
                  Bangbang Liu and
                  Tianqi Pang and
                  Jacky Kwok and
                  Ceyao Zhang and
                  Bang Liu and
                  Chenglin Wu},
  title        = {{SELA:} Tree-Search Enhanced {LLM} Agents for Automated Machine Learning},
  journal      = {CoRR},
  volume       = {abs/2410.17238},
  year         = {2024},
  url          = {https://doi.org/10.48550/arXiv.2410.17238},
  doi          = {10.48550/ARXIV.2410.17238},
  eprinttype   = {arXiv},
  eprint       = {2410.17238},
  bibsource    = {dblp computer science bibliography, https://dblp.org}
}

@article{shinn2023reflexion,
  title={Reflexion: Language agents with verbal reinforcement learning},
  author={Shinn, Noah and Cassano, Federico and Gopinath, Ashwin and Narasimhan, Karthik and Yao, Shunyu},
  journal={Advances in neural information processing systems},
  volume={36},
  pages={8634--8652},
  year={2023}
}

@article{madaan2023self,
  title={Self-refine: Iterative refinement with self-feedback},
  author={Madaan, Aman and Tandon, Niket and Gupta, Prakhar and Hallinan, Skyler and Gao, Luyu and Wiegreffe, Sarah and Alon, Uri and Dziri, Nouha and Prabhumoye, Shrimai and Yang, Yiming and others},
  journal={Advances in neural information processing systems},
  volume={36},
  pages={46534--46594},
  year={2023}
}

@article{DBLP:journals/tmlr/WangX0MXZFA24,
  author       = {Guanzhi Wang and
                  Yuqi Xie and
                  Yunfan Jiang and
                  Ajay Mandlekar and
                  Chaowei Xiao and
                  Yuke Zhu and
                  Linxi Fan and
                  Anima Anandkumar},
  title        = {Voyager: An Open-Ended Embodied Agent with Large Language Models},
  journal      = {Trans. Mach. Learn. Res.},
  volume       = {2024},
  year         = {2024},
  url          = {https://openreview.net/forum?id=ehfRiF0R3a},
  bibsource    = {dblp computer science bibliography, https://dblp.org}
}

@inproceedings{zhao2024expel,
  title={Expel: Llm agents are experiential learners},
  author={Zhao, Andrew and Huang, Daniel and Xu, Quentin and Lin, Matthieu and Liu, Yong-Jin and Huang, Gao},
  booktitle={Proceedings of the AAAI Conference on Artificial Intelligence},
  volume={38},
  number={17},
  pages={19632--19642},
  year={2024}
}

@article{DBLP:journals/pami/WangCLJHZLHZYML25,
  author       = {Zihao Wang and
                  Shaofei Cai and
                  Anji Liu and
                  Yonggang Jin and
                  Jinbing Hou and
                  Bowei Zhang and
                  Haowei Lin and
                  Zhaofeng He and
                  Zilong Zheng and
                  Yaodong Yang and
                  Xiaojian Ma and
                  Yitao Liang},
  title        = {{JARVIS-1:} Open-World Multi-Task Agents With Memory-Augmented Multimodal
                  Language Models},
  journal      = {{IEEE} Trans. Pattern Anal. Mach. Intell.},
  volume       = {47},
  number       = {3},
  pages        = {1894--1907},
  year         = {2025},
  url          = {https://doi.org/10.1109/TPAMI.2024.3511593},
  doi          = {10.1109/TPAMI.2024.3511593},
  bibsource    = {dblp computer science bibliography, https://dblp.org}
}

@inproceedings{guo2024ds,
  title={DS-agent: automated data science by empowering large language models with case-based reasoning},
  author={Guo, Siyuan and Deng, Cheng and Wen, Ying and Chen, Hechang and Chang, Yi and Wang, Jun},
  booktitle={Proceedings of the 41st International Conference on Machine Learning},
  pages={16813--16848},
  year={2024}
}

@inproceedings{zhang2024mlcopilot,
  title={Mlcopilot: Unleashing the power of large language models in solving machine learning tasks},
  author={Zhang, Lei and Zhang, Yuge and Ren, Kan and Li, Dongsheng and Yang, Yuqing},
  booktitle={Proceedings of the 18th Conference of the European Chapter of the Association for Computational Linguistics (Volume 1: Long Papers)},
  pages={2931--2959},
  year={2024}
}

@article{kim2026solve,
  title={Why Solve It Twice? Hierarchical Accumulation of Skills for Transfer-Efficient ML Engineering},
  author={Kim, Yongbin and Talebirad, Yashar and Zaiane, Osmar R},
  journal={arXiv preprint arXiv:2606.30911},
  year={2026}
}

@article{DBLP:journals/corr/abs-2602-08234,
  author       = {Peng Xia and
                  Jianwen Chen and
                  Hanyang Wang and
                  Jiaqi Liu and
                  Kaide Zeng and
                  Yu Wang and
                  Siwei Han and
                  Yiyang Zhou and
                  Xujiang Zhao and
                  Haifeng Chen and
                  Zeyu Zheng and
                  Cihang Xie and
                  Huaxiu Yao},
  title        = {SkillRL: Evolving Agents via Recursive Skill-Augmented Reinforcement
                  Learning},
  journal      = {CoRR},
  volume       = {abs/2602.08234},
  year         = {2026},
  url          = {https://doi.org/10.48550/arXiv.2602.08234},
  doi          = {10.48550/ARXIV.2602.08234},
  eprinttype   = {arXiv},
  eprint       = {2602.08234},
  bibsource    = {dblp computer science bibliography, https://dblp.org}
}

@article{hendrycks2021measuring,
  title={Measuring mathematical problem solving with the math dataset},
  author={Hendrycks, Dan and Burns, Collin and Kadavath, Saurav and Arora, Akul and Basart, Steven and Tang, Eric and Song, Dawn and Steinhardt, Jacob},
  journal={arXiv preprint arXiv:2103.03874},
  year={2021}
}

@inproceedings{lightman2024let,
  title={Let's verify step by step},
  author={Lightman, Hunter and Kosaraju, Vineet and Burda, Yuri and Edwards, Harrison and Baker, Bowen and Lee, Teddy and Leike, Jan and Schulman, John and Sutskever, Ilya and Cobbe, Karl},
  booktitle={International Conference on Learning Representations},
  volume={2024},
  pages={39578--39601},
  year={2024}
}

@article{DBLP:journals/bioinformatics/BakerSGAHSK16,
  author       = {Simon Baker and
                  Ilona Silins and
                  Yufan Guo and
                  Imran Ali and
                  Johan H{\"{o}}gberg and
                  Ulla Stenius and
                  Anna Korhonen},
  title        = {Automatic semantic classification of scientific literature according
                  to the hallmarks of cancer},
  journal      = {Bioinform.},
  volume       = {32},
  number       = {3},
  pages        = {432--440},
  year         = {2016},
  url          = {https://doi.org/10.1093/bioinformatics/btv585},
  doi          = {10.1093/BIOINFORMATICS/BTV585},
  bibsource    = {dblp computer science bibliography, https://dblp.org}
}

@article{DBLP:journals/corr/abs-2311-07911,
  author       = {Jeffrey Zhou and
                  Tianjian Lu and
                  Swaroop Mishra and
                  Siddhartha Brahma and
                  Sujoy Basu and
                  Yi Luan and
                  Denny Zhou and
                  Le Hou},
  title        = {Instruction-Following Evaluation for Large Language Models},
  journal      = {CoRR},
  volume       = {abs/2311.07911},
  year         = {2023},
  url          = {https://doi.org/10.48550/arXiv.2311.07911},
  doi          = {10.48550/ARXIV.2311.07911},
  eprinttype   = {arXiv},
  eprint       = {2311.07911},
  bibsource    = {dblp computer science bibliography, https://dblp.org}
}

@article{chen2021evaluating,
  title={Evaluating large language models trained on code},
  author={Chen, Mark and Tworek, Jerry and Jun, Heewoo and Yuan, Qiming and Pinto, Henrique Ponde De Oliveira and Kaplan, Jared and Edwards, Harri and Burda, Yuri and Joseph, Nicholas and Brockman, Greg and others},
  journal={arXiv preprint arXiv:2107.03374},
  year={2021}
}

@inproceedings{fei2024lawbench,
  title={Lawbench: Benchmarking legal knowledge of large language models},
  author={Fei, Zhiwei and Shen, Xiaoyu and Zhu, Dawei and Zhou, Fengzhe and Han, Zhuo and Huang, Alan and Zhang, Songyang and Chen, Kai and Yin, Zhixin and Shen, Zongwen and others},
  booktitle={Proceedings of the 2024 conference on empirical methods in natural language processing},
  pages={7933--7962},
  year={2024}
}

@article{DBLP:journals/corr/abs-2412-14642,
  author       = {Jiatong Li and
                  Junxian Li and
                  Yunqing Liu and
                  Dongzhan Zhou and
                  Qing Li},
  title        = {TOMG-Bench: Evaluating LLMs on Text-based Open Molecule Generation},
  journal      = {CoRR},
  volume       = {abs/2412.14642},
  year         = {2024},
  url          = {https://doi.org/10.48550/arXiv.2412.14642},
  doi          = {10.48550/ARXIV.2412.14642},
  eprinttype   = {arXiv},
  eprint       = {2412.14642},
  bibsource    = {dblp computer science bibliography, https://dblp.org}
}

@article{DBLP:journals/corr/abs-2507-02833,
  author       = {Valentina Pyatkin and
                  Saumya Malik and
                  Victoria Graf and
                  Hamish Ivison and
                  Shengyi Huang and
                  Pradeep Dasigi and
                  Nathan Lambert and
                  Hannaneh Hajishirzi},
  title        = {Generalizing Verifiable Instruction Following},
  journal      = {CoRR},
  volume       = {abs/2507.02833},
  year         = {2025},
  url          = {https://doi.org/10.48550/arXiv.2507.02833},
  doi          = {10.48550/ARXIV.2507.02833},
  eprinttype   = {arXiv},
  eprint       = {2507.02833},
  bibsource    = {dblp computer science bibliography, https://dblp.org}
}

@article{DBLP:journals/corr/abs-2311-12022,
  author       = {David Rein and
                  Betty Li Hou and
                  Asa Cooper Stickland and
                  Jackson Petty and
                  Richard Yuanzhe Pang and
                  Julien Dirani and
                  Julian Michael and
                  Samuel R. Bowman},
  title        = {{GPQA:} {A} Graduate-Level Google-Proof Q{\&}A Benchmark},
  journal      = {CoRR},
  volume       = {abs/2311.12022},
  year         = {2023},
  url          = {https://doi.org/10.48550/arXiv.2311.12022},
  doi          = {10.48550/ARXIV.2311.12022},
  eprinttype   = {arXiv},
  eprint       = {2311.12022},
  bibsource    = {dblp computer science bibliography, https://dblp.org}
}

@inproceedings{quan2024econlogicqa,
  title={Econlogicqa: A question-answering benchmark for evaluating large language models in economic sequential reasoning},
  author={Quan, Yinzhu and Liu, Zefang},
  booktitle={Findings of the Association for Computational Linguistics: EMNLP 2024},
  pages={2273--2282},
  year={2024}
}

@article{DBLP:journals/corr/abs-2306-02022,
  author       = {Wen{-}Wai Yim and
                  Yujuan Fu and
                  Asma Ben Abacha and
                  Neal Snider and
                  Thomas Lin and
                  Meliha Yetisgen},
  title        = {{ACI-BENCH:} a Novel Ambient Clinical Intelligence Dataset for Benchmarking
                  Automatic Visit Note Generation},
  journal      = {CoRR},
  volume       = {abs/2306.02022},
  year         = {2023},
  url          = {https://doi.org/10.48550/arXiv.2306.02022},
  doi          = {10.48550/ARXIV.2306.02022},
  eprinttype   = {arXiv},
  eprint       = {2306.02022},
  bibsource    = {dblp computer science bibliography, https://dblp.org}
}

@misc{openfindata2023,
    title        = "OpenFinData",
    author       = "{East Money} and {Shanghai AI Lab}",
    howpublished = "\url{https://github.com/open-compass/OpenFinData/}",
    year         = 2023,
    note         = "Accessed: 2026"
}

@article{zeng2026glm,
  title={Glm-5: from vibe coding to agentic engineering},
  author={Zeng, Aohan and Lv, Xin and Hou, Zhenyu and Du, Zhengxiao and Zheng, Qinkai and Chen, Bin and Yin, Da and Ge, Chendi and Huang, Chenghua and Xie, Chengxing and others},
  journal={arXiv preprint arXiv:2602.15763},
  year={2026}
}

@article{DBLP:journals/corr/abs-2412-15115,
  author       = {An Yang and
                  Baosong Yang and
                  Beichen Zhang and
                  Binyuan Hui and
                  Bo Zheng and
                  Bowen Yu and
                  Chengyuan Li and
                  Dayiheng Liu and
                  Fei Huang and
                  Haoran Wei and
                  Huan Lin and
                  Jian Yang and
                  Jianhong Tu and
                  Jianwei Zhang and
                  Jianxin Yang and
                  Jiaxi Yang and
                  Jingren Zhou and
                  Junyang Lin and
                  Kai Dang and
                  Keming Lu and
                  Keqin Bao and
                  Kexin Yang and
                  Le Yu and
                  Mei Li and
                  Mingfeng Xue and
                  Pei Zhang and
                  Qin Zhu and
                  Rui Men and
                  Runji Lin and
                  Tianhao Li and
                  Tingyu Xia and
                  Xingzhang Ren and
                  Xuancheng Ren and
                  Yang Fan and
                  Yang Su and
                  Yichang Zhang and
                  Yu Wan and
                  Yuqiong Liu and
                  Zeyu Cui and
                  Zhenru Zhang and
                  Zihan Qiu},
  title        = {Qwen2.5 Technical Report},
  journal      = {CoRR},
  volume       = {abs/2412.15115},
  year         = {2024},
  url          = {https://doi.org/10.48550/arXiv.2412.15115},
  doi          = {10.48550/ARXIV.2412.15115},
  eprinttype   = {arXiv},
  eprint       = {2412.15115},
  bibsource    = {dblp computer science bibliography, https://dblp.org}
}

@article{DBLP:journals/corr/abs-2409-12122,
  author       = {An Yang and
                  Beichen Zhang and
                  Binyuan Hui and
                  Bofei Gao and
                  Bowen Yu and
                  Chengpeng Li and
                  Dayiheng Liu and
                  Jianhong Tu and
                  Jingren Zhou and
                  Junyang Lin and
                  Keming Lu and
                  Mingfeng Xue and
                  Runji Lin and
                  Tianyu Liu and
                  Xingzhang Ren and
                  Zhenru Zhang},
  title        = {Qwen2.5-Math Technical Report: Toward Mathematical Expert Model via
                  Self-Improvement},
  journal      = {CoRR},
  volume       = {abs/2409.12122},
  year         = {2024},
  url          = {https://doi.org/10.48550/arXiv.2409.12122},
  doi          = {10.48550/ARXIV.2409.12122},
  eprinttype   = {arXiv},
  eprint       = {2409.12122},
  bibsource    = {dblp computer science bibliography, https://dblp.org}
}

@article{ball1997problem,
  title={Problem-solving strategies and expertise in engineering design},
  author={Ball, Linden J and St. BT Evans, Jonathan and Dennis, Ian and Ormerod, Thomas C},
  journal={Thinking \& Reasoning},
  volume={3},
  number={4},
  pages={247--270},
  year={1997},
  publisher={Taylor \& Francis}
}

@inproceedings{arab2022exploratory,
  title={An exploratory study of sharing strategic programming knowledge},
  author={Arab, Maryam and LaToza, Thomas D and Liang, Jenny and Ko, Amy J},
  booktitle={Proceedings of the 2022 CHI conference on human factors in computing systems},
  pages={1--15},
  year={2022}
}

@article{costa2022no,
  title={No language left behind: Scaling human-centered machine translation},
  author={Costa-Juss{\`a}, Marta R and Cross, James and {\c{C}}elebi, Onur and Elbayad, Maha and Heafield, Kenneth and Heffernan, Kevin and Kalbassi, Elahe and Lam, Janice and Licht, Daniel and Maillard, Jean and others},
  journal={arXiv preprint arXiv:2207.04672},
  year={2022}
}

@inproceedings{deutsch2025wmt24++,
  title={WMT24++: Expanding the language coverage of WMT24 to 55 languages \& dialects},
  author={Deutsch, Daniel and Briakou, Eleftheria and Caswell, Isaac Rayburn and Finkelstein, Mara and Galor, Rebecca and Juraska, Juraj and Kovacs, Geza and Lui, Alison and Rei, Ricardo and Riesa, Jason and others},
  booktitle={Findings of the Association for Computational Linguistics: ACL 2025},
  pages={12257--12284},
  year={2025}
}

@misc{zhao2024swiftascalablelightweightinfrastructure,
      title={SWIFT:A Scalable lightWeight Infrastructure for Fine-Tuning},
      author={Yuze Zhao and Jintao Huang and Jinghan Hu and Xingjun Wang and Yunlin Mao and Daoze Zhang and Zeyinzi Jiang and Zhikai Wu and Baole Ai and Ang Wang and Wenmeng Zhou and Yingda Chen},
      year={2024},
      eprint={2408.05517},
      archivePrefix={arXiv},
      primaryClass={cs.CL},
      url={https://arxiv.org/abs/2408.05517},
}

\clearpage

\appendix

\begin{center}
    {\LARGE\bfseries Supplementary Material\par}
\end{center}

\section{Limitations}
\label{app:limitations}

While SAGE has demonstrated the effectiveness of accumulating and reusing structured experience for automated LLM fine-tuning, several limitations remain:

\begin{itemize}

\item \textbf{Scale of exploration.} Our Stage~1 exploration is limited to three meta-tasks with 20 expansions each. Scaling to more meta-tasks per category or conducting deeper searches could further enrich the experience repository, but would require proportionally greater compute investment. The relationship between exploration budget and downstream Stage~2 performance remains to be characterized.

\item \textbf{Model scale.} Due to the substantial time cost of each MCTS node at larger model sizes, all experiments use 1.5B-parameter dense models as $\theta_0$. Validating SAGE at larger scales (7B, 70B) where the optimal training configurations may differ significantly remains an open question and a valuable direction for future work.

\item \textbf{Task diversity.} Our evaluation covers nine Stage~2 tasks in the explored categories, plus two out-of-category translation evaluations. Although these tasks provide coverage across several task families, fundamentally different training paradigms, such as multimodal grounding, tool use, or long-context specialization, have not been tested. These may require dedicated Stage~1 exploration to populate category-specific experience before SAGE can provide effective guidance.

\end{itemize}

Future work could address these limitations by (i) scaling Stage~1 exploration to more meta-tasks and deeper search trees to establish diminishing-return curves, (ii) validating experience transfer at larger model scales, and (iii) extending the category taxonomy to cover a broader range of capability bottlenecks and task families.

\section{Search and Confidence Hyperparameters}
\label{app:hyperparameters}

Tables~\ref{tab:mcts_hparams} and~\ref{tab:confidence_hparams} list the hyperparameters used in MCTS search and the confidence lifecycle for cross-task insight management, respectively.

\begin{table*}[ht]
  \centering
  \begin{tabular}{@{}lr@{}}
    \toprule
    Hyperparameter & Value \\
    \midrule
    UCT exploration constant $c$ & $\sqrt{2} \approx 1.414$ \\
    Progressive widening cap $k_{\max}$ & 4 \\
    Maximum tree depth $d_{\max}$ & 8 \\
    Completed expansions per search & 20 \\
    PaSR parent weight $\alpha$ & 0.3 \\
    GPU efficiency lower bound $e_{\min}$ & 0.5 \\
    Exploit threshold ($w <$) & 0.2 \\
    Explore threshold ($w >$) & 0.6 \\
    \bottomrule
  \end{tabular}
  \captionsetup{width=\textwidth,justification=centering,singlelinecheck=true}
  \caption{MCTS search hyperparameters.}
  \label{tab:mcts_hparams}
\end{table*}

\begin{table*}[ht]
  \centering
  \begin{tabular}{@{}lr@{}}
    \toprule
    Parameter & Value \\
    \midrule
    Initial confidence $c_\text{init}$ & $0.6$ \\
    Confidence step $\delta$ & $0.1$ \\
    Harmful decay factor $\gamma$ & $0.5$ \\
    Pruning threshold & $0.2$ \\
    \bottomrule
  \end{tabular}
  \captionsetup{width=\textwidth,justification=centering,singlelinecheck=true}
  \caption{Cross-task insight confidence lifecycle constants.}
  \label{tab:confidence_hparams}
\end{table*}

\FloatBarrier

\section{Algorithms}
\label{app:algorithm}

\subsection{MCTS-Based Strategy Exploration}

Algorithm~\ref{alg:mcts} makes the two-level structure of Stage~1 explicit: MCTS selects a parent node, and the role-isolated multi-agent pipeline performs one expansion to produce a child node. Exact node scores remain internal to the orchestrator and Reviewer; subsequent Planner calls receive qualitative tree context and score-free critiques.

\begin{algorithm}[H]
\caption{MCTS-Based Multi-Agent Strategy Exploration}
\label{alg:mcts}
\small
\begin{algorithmic}[1]
\REQUIRE Task $\tau$, base model $\theta_0$, budget $N_{\max}$, caps $k_{\max},d_{\max}$
\ENSURE Search tree $\mathcal{T}$, experience update $\Delta\mathcal{E}(\tau)$
\STATE $s_{\mathrm{base}} \leftarrow \mathrm{eval}_\tau(\theta_0)$; initialize root $v_0$ with $s(v_0)=s_{\max}=s_{\mathrm{base}}$ and $Q(v_0)=N(v_0)=0$
\STATE $\mathcal{T} \leftarrow \{v_0\}$; $n \leftarrow 0$
\WHILE{$n < N_{\max}$}
    \STATE \textbf{// 1. Selection}
    \STATE $\mathcal{C} \leftarrow$ scored nodes in $\mathcal{T}$ below the child and depth caps
    \IF{$\mathcal{C}=\emptyset$} \STATE \textbf{break} \ENDIF
    \STATE $v \leftarrow v_0$ if $n=0$; otherwise $v \leftarrow \arg\max_{u\in\mathcal{C}}\mathrm{UCT}(u)$
    \STATE $q \leftarrow 0$ if $s_{\max}\le s_{\mathrm{base}}$; otherwise $\mathrm{clip}\!\left(\frac{s(v)-s_{\mathrm{base}}}{s_{\max}-s_{\mathrm{base}}},0,1\right)$
    \STATE $w \leftarrow \mathrm{clip}\!\left(1-q-\min(0.1\,\mathrm{depth}(v),0.3),0,1\right)$
    \STATE \textbf{// 2. Expansion via Planner and Executor}
    \STATE $\mathrm{plan} \leftarrow \mathrm{Planner}(\tau,w,\mathrm{AgentView}(v,\mathcal{T}))$
    \STATE $\theta' \leftarrow \mathrm{Executor}(\theta_0,\mathrm{plan})$
    \STATE \textbf{// 3. Evaluation and Review (replacing a cheap rollout)}
    \STATE $s' \leftarrow \mathrm{OrchestratorEval}_\tau(\theta')$ \hfill $\triangleright$ retained by the orchestrator
    \STATE $\mathrm{fair} \leftarrow \mathrm{Reviewer}(\mathrm{plan},\theta',s')$
    \IF{$\neg\mathrm{fair}$}
        \STATE Record abandoned attempt; \textbf{continue}
    \ENDIF
    \STATE $\mathrm{critique} \leftarrow \mathrm{Critic}(\mathrm{plan},\mathrm{Predictions}(\theta'))$ \hfill $\triangleright$ score-free report
    \STATE Create child $v'$ of $v$ from $(\mathrm{plan},\theta',s',\mathrm{critique})$ with $Q(v')=N(v')=0$; add $v'$ to $\mathcal{T}$
    \STATE \textbf{// 4. PaSR Backpropagation and Distillation}
    \STATE $r \leftarrow [(s'-s_{\mathrm{base}})+\alpha(s'-s(v))]\,e(v')$ \hfill $\triangleright$ PaSR
    \STATE \textsc{Backpropagate}$(v',r)$; $s_{\max}\leftarrow\max(s_{\max},s')$; $n\leftarrow n+1$
    \STATE Launch node-level distillation for $v'$ asynchronously
\ENDWHILE
\STATE $\Delta\mathcal{E}(\tau) \leftarrow$ chain-level and whole-tree distillation of $\mathcal{T}$
\STATE \textbf{return} $\mathcal{T},\Delta\mathcal{E}(\tau)$
\end{algorithmic}
\end{algorithm}

\paragraph{Operational semantics.} $N_{\max}$ counts reviewer-approved expansions, so an abandoned attempt creates no scored child and does not increment $n$. While $n=0$, selection uses $v_0$ directly. For an accepted child, \textsc{Backpropagate} adds $r$ to $Q(u)$ and one to $N(u)$ along the child-to-root path, so every later candidate has $N(u)>0$ and the parentless root uses $Q(v_0)/N(v_0)$. Candidate nodes are otherwise compared globally by UCT; \textsc{AgentView} provides qualitative tree history and sibling summaries without exact scores. PaSR then combines improvement over the baseline and parent, while node-, chain-, and tree-level distillation extract experience at progressively broader scopes.

\subsection{Experience Retrieval and Single-Round Execution}

Algorithm~\ref{alg:stage2} specifies one Stage~2 execution round. The Planner retrieves records for every matched category, compares the target and source settings through gap analysis, and checks each insight's transfer and break conditions before using it. From Round~2 onward, it also receives the preceding plan and score-free Critic report. The reported Stage~2 evaluation reads a frozen repository and does not update the accumulated experience.

\begin{algorithm}[H]
\caption{Experience Retrieval and Single-Round Execution}
\label{alg:stage2}
\small
\begin{algorithmic}[1]
\REQUIRE New task $\tau'$, base model $\theta_0$, frozen $\mathcal{E}$, prior plan $p_{\mathrm{prev}}$ and score-free critique $c_{\mathrm{prev}}$ (both empty in Round~1)
\ENSURE $(\theta',p,c)$, or an abandoned status
\STATE $s_{\text{base}} \leftarrow \mathrm{eval}_{\tau'}(\theta_0)$
\STATE \textbf{// 1. Category-Matched Records and Global Insights}
\STATE $\mathcal{R} \leftarrow \textsc{RetrieveRecordsByCategory}(\mathcal{E},g(\tau'))$
\STATE $\mathcal{I} \leftarrow \textsc{AllCrossTaskInsights}(\mathcal{E})$
\STATE \textbf{// 2. Planner-Side Applicability Checking}
\STATE $p \leftarrow \mathrm{Planner}(\tau',s_{\mathrm{base}},\mathcal{R},\mathcal{I},p_{\mathrm{prev}},c_{\mathrm{prev}})$
\STATE \hfill $\triangleright$ compare source/target baselines and evaluation mechanisms
\STATE \hfill $\triangleright$ verify \textit{transfer-when} and \textit{transfer-breaks} before adoption
\STATE \textbf{// 3. Single-Round Multi-Agent Execution}
\STATE $\theta' \leftarrow \mathrm{Executor}(\theta_0,p)$ \hfill $\triangleright$ always restart from $\theta_0$
\STATE $s' \leftarrow \mathrm{OrchestratorEval}_{\tau'}(\theta')$
\STATE $\mathrm{fair} \leftarrow \mathrm{Reviewer}(p,\theta',s')$
\IF{$\neg\mathrm{fair}$}
    \STATE \textbf{return} abandoned
\ENDIF
\STATE $c \leftarrow \mathrm{Critic}(p,\mathrm{Predictions}(\theta'))$
\STATE \textbf{return} $\theta',p,c$
\end{algorithmic}
\end{algorithm}

Unlike Stage~1, this procedure performs no tree search. It retrieves accumulated experience, lets the Planner assess transferability, and executes the multi-agent pipeline once. Algorithm~\ref{alg:stage2} specifies the single-round execution unit; the multi-round evaluation repeats this unit while threading only $(p,c)$ forward. Exact scores remain with the orchestrator, model weights are discarded, and the next plan is applied afresh to $\theta_0$. The repository remains frozen throughout the reported Stage~2 evaluation so every run starts from the same Stage~1 experience.

\begin{table*}[t]
\centering
{%
\small
\begin{tabular}{@{}lllllr@{}}
\toprule
Stage & Task & Category & Base Model & Metric & Eval Size \\
\midrule
1 & MATH500 & R & Qwen2.5-Math-1.5B-Instruct & Accuracy & 500 \\
1 & HoC & K & Qwen2.5-1.5B-Instruct & Macro-F1 & 315 \\
1 & IFEval & A & Qwen2.5-1.5B-Instruct & Prompt-level strict acc. & 541 \\
\midrule
2 & AMC23 & R & Qwen2.5-Math-1.5B-Instruct & Accuracy & 40 \\
2 & HumanEval & R & Qwen2.5-Math-1.5B-Instruct & Pass@1 & 164 \\
2 & LawBench & K & Qwen2.5-1.5B-Instruct & Avg.\ Score (multi-task) & 5,000 \\
2 & ToMG-Bench & K & Qwen2.5-1.5B-Instruct & Validity \& Accuracy & 2,200 \\
2 & OpenFindata & K & Qwen2.5-1.5B-Instruct & Accuracy & 650 \\
2 & IFBench & A & Qwen2.5-1.5B-Instruct & Acc.$_{\text{prompt-loose}}$ & 300 \\
2 & GPQA-Diamond & R+K & Qwen2.5-Math-1.5B-Instruct & Accuracy & 198 \\
2 & EconLogicQA & R+K & Qwen2.5-Math-1.5B-Instruct & Accuracy & 130 \\
2 & ACI-Bench & K+A & Qwen2.5-1.5B-Instruct & ROUGE-1 & 600 \\
\midrule
OOD & WMT24PP & N/A & Qwen2.5-1.5B-Instruct & SacreBLEU & 960 \\
OOD & FLORES & N/A & Qwen2.5-1.5B-Instruct & SacreBLEU & 1,012 \\
\multicolumn{6}{@{}l@{}}{\footnotesize\textit{Both OOD evaluations use the English-to-Simplified-Chinese (En$\to$Zh-CN) split.}} \\
\bottomrule
\end{tabular}
}
\caption{Task specifications for Stage~1 meta-tasks, Stage~2 evaluation tasks, and OOD evaluations, including each task's category, base model, evaluation metric, and evaluation-set size. Stage~1 tasks support exploration and experience accumulation, whereas Stage~2 and OOD tasks assess transfer to previously unseen tasks and distribution shifts. Category abbreviations: R=Reasoning, K=Knowledge, and A=Alignment.}
\label{tab:task_specs}
\end{table*}

\section{Task Specifications}
\label{app:tasks}

Table~\ref{tab:task_specs} summarizes the evaluation tasks used across both stages, including category assignments, evaluation metrics, and the number of examples scored in each run.

\begin{table*}[t]
\centering
{%
\small
\setlength{\tabcolsep}{3.5pt}
\begin{tabular}{@{}lllccl@{}}
\toprule
\textit{Node} &
\textit{Key Change} &
\textit{Configuration} &
\textit{Score} &
\textit{$\Delta$Parent} &
\textit{Observation} \\
\midrule
root &
baseline &
Qwen2.5-Math-1.5B-Instruct &
0.366 &
--- &
--- \\

root\_2 &
+SFT+GRPO &
SFT 3ep b8 ml=4096; GRPO lora64 lr=1e-5 &
0.682 &
+0.316 &
largest path gain \\

root\_2\_1 &
cleaned data; GRPO lora$\to$full &
SFT 5ep ml=2048 no-gc; GRPO full lr=2e-5 &
0.716 &
+0.034 &
additional gain \\

root\_2\_1\_2 &
data 16K$\to$125K; ml$\downarrow$ &
SFT 3ep b20 ml=1024 gc; GRPO kl=0.1 &
0.676 &
$-$0.040 &
bundle regresses \\

\textbf{root\_2\_1\_2\_5} &
\textbf{quality-filtered 80K; ml$\uparrow$; no-gc} &
\textbf{SFT 2ep b8 ml=2048; GRPO 120st} &
\textbf{0.732} &
\textbf{+0.056} &
\textbf{best path score} \\
\bottomrule
\end{tabular}
}
\captionsetup{justification=raggedright,singlelinecheck=false}
\caption{Representative MATH500 exploration trajectory from the root to the best-performing node in the search tree. Each row summarizes the principal strategy change at one depth, illustrating how MCTS discovers the strongest explored strategy while SAGE accumulates the intermediate changes and outcomes as reusable experience.}
\label{tab:training_record}
\end{table*}

For cross-category tasks (GPQA-Diamond, EconLogicQA, ACI-Bench), the Planner retrieves experience from all matched categories. For out-of-category tasks (WMT24PP, FLORES), no exploration records exist; the Planner draws only on cross-task insights whose applicability conditions are satisfied.

\section{Data Isolation and Contamination Control}
\label{app:isolation}

Because SAGE uses LLM agents that can autonomously select and construct training data, rigorous isolation between training and evaluation is essential. We enforce the following safeguards:

\paragraph{Sandbox enforcement.} Each agent operates within a restricted filesystem sandbox. The Executor has read-write access only to its designated node directory (training data, outputs, and logs). Test data and gold labels are not mounted in the Executor sandbox and are therefore inaccessible. The evaluation script and other non-sensitive context are mounted read-only so the Executor can align its output format with the deterministic evaluation interface without accessing evaluation examples. The sandbox is enforced via bind-mount permissions configured before each agent launch.

\paragraph{Compliance checking.} After each Executor run, an automated checker performs conservative, deterministic screening for directly verifiable violations. It checks for apparent access to protected test files, training invocations beyond the stages declared in the plan, exact matches between protected identifiers or inputs and available structured training records, and missing or empty merged-model outputs. Read-only inspection of the evaluation script is permitted, and ambiguous trace patterns are not classified as violations. These checks therefore serve as guardrails rather than a complete proof of data isolation. Nodes with confirmed violations are marked as abandoned and excluded from valid expansions. The Distillation Agent may record the failure type, but it does not treat the score or strategy outcome of an abandoned node as evidence about strategy effectiveness.

\paragraph{Reviewer verification.} The Reviewer agent complements the deterministic checker with a contextual audit of each completed node. It verifies that no test data or gold answers were used for training, that the executed training stages agree with the declared plan without unplanned restarts, and that resource usage is reported consistently with the trace. Confirmed fairness violations trigger node abandonment.

\begin{table*}[!t]
\centering
{%
\small
\setlength{\tabcolsep}{3pt}
\begin{tabular}{@{}llcccc@{}}
\toprule
Task & Condition & Independent R1 runs & Avg@3 & Sequential rounds (R1/R2/R3) & Best@3R \\
\midrule
AMC23 & w/o SAGE & $0.325\,/\,0.175\,/\,0.250$ & 0.250 & $0.325\,/\,0.250\,/\,0.200$ & 0.325 \\
& w/ SAGE & $0.425\,/\,0.475\,/\,0.375$ & 0.425 & $0.475\,/\,0.325\,/\,0.375$ & 0.475 \\
\midrule
HumanEval & w/o SAGE & $0.384\,/\,0.311\,/\,0.006$ & 0.234 & $0.006\,/\,0.348\,/\,0.354$ & 0.354 \\
& w/ SAGE & $0.427\,/\,0.378\,/\,0.256$ & 0.354 & $0.378\,/\,0.396\,/\,0.415$ & 0.415 \\
\midrule
LawBench & w/o SAGE & $0.407\,/\,0.270\,/\,0.332$ & 0.336 & $0.407\,/\,0.378\,/\,0.399$ & 0.407 \\
& w/ SAGE & $0.388\,/\,0.348\,/\,0.462$ & 0.399 & $0.388\,/\,0.523\,/\,0.341$ & 0.523 \\
\midrule
ToMG-Bench & w/o SAGE & $0.422\,/\,0.344\,/\,0.321$ & 0.362 & $0.344\,/\,0.104\,/\,0.491$ & 0.491 \\
& w/ SAGE & $0.453\,/\,0.417\,/\,0.339$ & 0.403 & $0.453\,/\,0.660\,/\,0.409$ & 0.660 \\
\midrule
OpenFindata & w/o SAGE & $0.585\,/\,0.532\,/\,0.554$ & 0.557 & $0.585\,/\,0.492\,/\,0.500$ & 0.585 \\
& w/ SAGE & $0.651\,/\,0.622\,/\,0.645$ & 0.639 & $0.651\,/\,0.580\,/\,0.608$ & 0.651 \\
\midrule
IFBench & w/o SAGE & $0.153\,/\,0.193\,/\,0.147$ & 0.164 & $0.147\,/\,0.220\,/\,0.240$ & 0.240 \\
& w/ SAGE & $0.290\,/\,0.250\,/\,0.193$ & 0.244 & $0.290\,/\,0.233\,/\,0.420$ & 0.420 \\
\midrule
GPQA-Diamond & w/o SAGE & $0.131\,/\,0.147\,/\,0.232$ & 0.170 & $0.232\,/\,0.253\,/\,0.081$ & 0.253 \\
& w/ SAGE & $0.237\,/\,0.298\,/\,0.242$ & 0.259 & $0.237\,/\,0.288\,/\,0.308$ & 0.308 \\
\midrule
EconLogicQA & w/o SAGE & $0.108\,/\,0.023\,/\,0.015$ & 0.049 & $0.015\,/\,0.000\,/\,0.108$ & 0.108 \\
& w/ SAGE & $0.177\,/\,0.100\,/\,0.177$ & 0.151 & $0.177\,/\,0.108\,/\,0.154$ & 0.177 \\
\midrule
ACI-Bench & w/o SAGE & $0.274\,/\,0.201\,/\,0.172$ & 0.215 & $0.201\,/\,0.362\,/\,0.162$ & 0.362 \\
& w/ SAGE & $0.348\,/\,0.257\,/\,0.278$ & 0.294 & $0.348\,/\,0.334\,/\,0.493$ & 0.493 \\
\bottomrule
\end{tabular}
}
\captionsetup{justification=raggedright,singlelinecheck=false}
\caption{Per-run results on the nine Stage~2 tasks. Slashes preserve execution order. ACI-Bench ROUGE-1 scores are divided by 100 to match the 0--1 scale used in the main-results table. All other tasks use the metrics and scales listed in Table~\ref{tab:task_specs}.}
\label{tab:stage2_per_run}
\end{table*}

\paragraph{Score isolation.} During strategy execution, the Planner and Executor do not receive the exact score of the current node or round. Deterministic evaluation returns that score to the orchestrator, which retains it for UCT selection, PaSR backpropagation, best-model selection, and aggregate reporting, and also provides it to the Reviewer for outcome assessment. The Critic may inspect prediction artifacts for bad-case analysis but is instructed not to read or reproduce aggregate benchmark scores; subsequent Planner calls receive its score-free diagnosis and qualitative outcome summaries. In Stage~2, the Planner can read source-task baselines and node scores stored in exploration records, and it receives the target task's baseline for the numerical gap analysis defined in the main paper. It still cannot access the exact scores of preceding Stage~2 training rounds.

\paragraph{Cross-stage isolation.} Node-level distillation runs asynchronously after each completed expansion; chain-level and whole-tree consolidation run after the search finishes. The Stage~2 experience view contains strategy descriptions, distilled patterns, source-task scores, and applicability conditions, but never raw test samples, gold labels, or per-sample predictions. Stage~2 tasks are entirely unseen during Stage~1: no Stage~2 task names, test sets, or evaluation criteria appear in the Stage~1 pipeline.

In the MATH500 exploration, 6 out of 26 total expansion attempts were abandoned due to compliance violations (all cases of unauthorized repeated training), demonstrating that the safeguards actively enforce experimental integrity.

\section{Detailed Per-Run Results}
\label{app:per_run_results}

Tables~\ref{tab:stage2_per_run} and~\ref{tab:ood_per_run} expand the aggregate results reported in the main paper. The three entries under ``Independent R1 runs'' are separate Round~1 executions whose unrounded mean gives Avg@3. The three entries under ``Sequential rounds'' belong to one refinement chain that carries the prior plan and score-free critique but restarts each Executor run from $\theta_0$; their maximum gives Best@3R. All entries are rounded to three decimal places for display; aggregate values are computed from unrounded scores.

\begin{center}
\begin{minipage}{\textwidth}
\centering
{%
\small
\setlength{\tabcolsep}{5pt}
\begin{tabular}{@{}llcccc@{}}
\toprule
Task & Condition & Independent R1 runs & Avg@3 & Sequential rounds (R1/R2/R3) & Best@3R \\
\midrule
WMT24PP & w/o SAGE & $0.245\,/\,0.284\,/\,0.290$ & 0.273 & $0.290\,/\,0.276\,/\,0.277$ & 0.290 \\
& w/ SAGE & $0.312\,/\,0.307\,/\,0.305$ & 0.308 & $0.312\,/\,0.328\,/\,0.351$ & 0.351 \\
\midrule
FLORES & w/o SAGE & $0.185\,/\,0.336\,/\,0.336$ & 0.286 & $0.336\,/\,0.348\,/\,0.351$ & 0.351 \\
& w/ SAGE & $0.396\,/\,0.372\,/\,0.372$ & 0.380 & $0.396\,/\,0.400\,/\,0.406$ & 0.406 \\
\bottomrule
\end{tabular}
}
\captionsetup{hypcap=false,justification=raggedright,singlelinecheck=false}
\captionof{table}{Per-run out-of-category results on the En$\to$Zh-CN splits of WMT24PP and FLORES. SacreBLEU scores are divided by 100 for a consistent 0--1 display scale, and slashes preserve execution order.}
\label{tab:ood_per_run}
\end{minipage}
\end{center}

Taken together, these supplementary materials provide a consolidated account of SAGE's two-stage workflow, from strategy search and safeguarded execution to experience representation and cross-task reuse.

\section{Experience Repository Examples}
\label{app:experience_examples}

This section provides concrete examples from the accumulated experience repository to illustrate the two-tier structure described in the Experience Accumulation section of the main paper. Exploration-record scores remain available to the Stage~2 Planner for source--target gap analysis, whereas raw evaluation examples and predictions are never included in the repository. We show a representative MATH500 evolution trajectory, two cross-task insights, and the fixed schema shared by both tiers.

\subsection{Exploration Record: Representative Evolution Trajectory}
\label{app:record_training}

Rather than summarizing heterogeneous trials by training method, Table~\ref{tab:training_record} replays the root-to-best trajectory from the MATH500 exploration. Each row isolates the principal strategy change made at one depth, making the evolution of the final training recipe explicit.

\FloatBarrier

\subsection{Cross-Task Insight Example: Method Dimension}
\label{app:insight_example_method}

The following is a representative high-confidence insight from the method dimension, accumulated during MATH500 and IFEval exploration.

\begin{insightbox}{Limited GRPO Gains after Reward Saturation \hfill \normalfont\itshape confidence = 1.4}
\small
\textit{Dimension:} method \quad \textit{Category:} reasoning\\[-1pt]
\textit{Origin:} MATH500 (baseline = 0.366)

\textbf{Statement.} In the recorded runs, GRPO often added little after SFT had already achieved high reward on a closely aligned signal. This pattern coincided with high reward, low loss, or limited reward sensitivity and should be treated as an empirical heuristic rather than a general claim about GRPO.

\textbf{Mechanism.} One plausible explanation is reduced advantage variance when generations and rewards become less diverse. The repository records this as a diagnostic hypothesis, not a verified causal mechanism.

\textbf{Transfers when:} SFT and GRPO use closely aligned rewards and the SFT run already appears saturated.

\textbf{Breaks when:} GRPO introduces a substantially different or denser reward, valid outputs remain diverse, or the SFT reward is not saturated.

\tcblower
\footnotesize
\textbf{Validation Records} (representative subset; validated across MATH500 and IFEval):\\[2pt]
\begin{tabular}{@{}lll@{}}
\textit{Task} & \textit{Node} & \textit{Result} \\
MATH500 & root\_2\_2\_2 & confirmed \\
IFEval & root\_1\_1\_2 & confirmed \\
\multicolumn{3}{@{}l@{}}{\textit{\ldots (additional records omitted for brevity)}}
\end{tabular}
\end{insightbox}

This insight informed the IFBench case study in the main paper. The Planner's gap analysis found substantially more headroom on IFBench than in the IFEval origin setting. The Planner therefore included GRPO with a custom constraint-verification reward because IFBench satisfied the insight's \textit{breaks-when} condition for a substantially different reward.

\subsection{Cross-Task Insight Example: Dataset Dimension}
\label{app:insight_example_dataset}

A second representative insight from the dataset dimension, accumulated during IFEval exploration:

\begin{insightbox}{Unverified Data Can Degrade Constraint Learning \hfill \normalfont\itshape confidence = 1.4}
\small
\textit{Dimension:} dataset \quad \textit{Category:} alignment\\[-1pt]
\textit{Origin:} IFEval (baseline = 0.388)

\textbf{Statement.} Across the recorded IFEval runs, unverified SFT data was associated with weaker constraint learning, particularly when violations or repeated patterns were over-represented. This is an empirical observation and may not transfer unchanged to other tasks.

\textbf{Mechanism.} A plausible explanation is that SFT tracks the observed data distribution and can amplify systematic patterns already present in the data. The recorded density thresholds are therefore run-specific diagnostics rather than universal cutoffs.

\textbf{Transfers when:} Scraped or generated data contains measurable constraint violations, repeated patterns, or systematic artifacts.

\textbf{Breaks when:} Explicit negative supervision is available, the suspect pattern is rare, or the base model has little prior tendency toward it.

\tcblower
\footnotesize
\textbf{Origin-Task Validation Records} (representative subset from IFEval):\\[2pt]
\begin{tabular}{@{}lll@{}}
\textit{Task} & \textit{Node} & \textit{Result} \\
IFEval & root\_2 & confirmed \\
IFEval & root\_2\_2\_2 & confirmed \\
\multicolumn{3}{@{}l@{}}{\textit{\ldots (additional records omitted for brevity)}}
\end{tabular}
\end{insightbox}

In Stage~2, this insight led the IFBench Planner to programmatically verify training data before use. The resulting run improved by 30.1\% relative to baseline, although the available evidence does not isolate data verification as the sole cause. In the w/o~SAGE run, the Planner selected instruction-following datasets whose constraint types did not overlap with IFBench's 58 novel test types.

\Needspace{0.50\textheight}
\subsection{Experience Schema and Traceability}
\label{app:experience_schema}

SAGE represents reusable experience through the two conceptual tiers introduced in the main paper. Fixed schemas make their experimental evidence comparable, retrievable, and auditable rather than an undifferentiated collection of free-form notes.

\paragraph{Tier 1: Exploration records.}
Each explored task contains \texttt{overview.md}, \texttt{training.md}, and \texttt{data.md}. The overview combines task metadata (baseline, best node and score, and node count) with a fixed node table: parent, changed dimension, method (M), dataset (D), format (F), hyperparameters (H), score, baseline and parent deltas, and assessment. This M/D/F/H decomposition makes tree edits directly comparable. The other files reorganize node outcomes into dimension-specific rules and comparison tables, with explicit transfer boundaries.

\paragraph{Tier 2: Cross-task insights.}
Each insight belongs to one of four dimensions (method, dataset, format, or hyperparameters) and records confidence, task category, statement, mechanism hypothesis, transfer and break conditions, and origin. Its validation table traces confirmations, revisions, and contradictions to concrete tasks and nodes, with measured deltas and evidence. Wiki-style links connect related insights into an evidence graph.

\paragraph{Why the representation is structured.}
The schema separates empirical claims from mechanism hypotheses and applicability boundaries. Enumerated dimensions and task categories support precise retrieval. Confidence updates follow the lifecycle defined in the main paper, and node-level provenance makes each update auditable. The Stage~2 Planner can therefore check explicit fields instead of inferring all relevant information from free-form narratives.

\end{document}